\pdfoutput=1
\documentclass{article}
\PassOptionsToPackage{table}{xcolor}
\usepackage{iclr2027_conference,times}
\iclrfinalcopy
\usepackage[T1]{fontenc}
\usepackage[utf8]{inputenc}
\usepackage{microtype,amsmath,amssymb,graphicx,booktabs,tabularx,multirow,xcolor}
\usepackage{tikz}
\usepackage{siunitx}
\usepackage{array}
\usetikzlibrary{arrows.meta,positioning,fit,calc}
\usepackage{xurl}
\usepackage[colorlinks=true,citecolor=blue!55!black,urlcolor=blue!55!black,linkcolor=blue!55!black]{hyperref}
\hypersetup{pdftitle={DeskForge: Dense Supervision from Desktop Environments for Computer-Use Agents},
  pdfauthor={A. Said Gurbuz, Ahmed Nassar, Sunghwan Hong, Marc Pollefeys, Peter W. J. Staar}}
\definecolor{resultbest}{RGB}{255,232,205}
\definecolor{resultsecond}{RGB}{255,245,232}
\newcommand{\framework}{DeskForge}
\newcommand{\dataset}{DeskForge-1M}
\newcommand{\code}[1]{\texttt{\detokenize{#1}}}
\newcommand{\cmark}{{\color{green!50!black}\checkmark}}
\newcommand{\xmark}{{\color{red!70!black}$\times$}}

\title{DeskForge: Dense Supervision From Desktop \\ Environments for Computer-use agents}
\author{\parbox[t]{\dimexpr\textwidth-2\tabcolsep\relax}{\centering
A. Said Gurbuz\textsuperscript{1,2} \quad Ahmed Nassar\textsuperscript{2} \quad Sunghwan Hong\textsuperscript{1} \\
Marc Pollefeys\textsuperscript{1,3} \quad Peter W. J. Staar\textsuperscript{2} \\[2pt]
\textnormal{\textsuperscript{1}ETH Zurich \quad \textsuperscript{2}IBM Research Zurich \quad \textsuperscript{3}Microsoft} \\[2pt]
\textnormal{\small Project page: \url{https://saidgurbuz.github.io/deskforge/}}}}
\date{}
\begin{document}
\maketitle
\lhead{Preprint}
\begin{abstract}
Computer-use agents need to reliably ground action targets in complex desktop scenes, where multiple applications, overlapping windows, and visually similar controls compete for attention. Existing training data rarely pair such scenes with dense annotations or vary them in a controlled way. We introduce \textbf{\framework}, a controllable desktop environment that composes and explores real applications to generate large-scale supervision for computer-use agents. It varies application states, content, window layout, appearance, and resolution, and fuses screenshots, accessibility trees, and window geometry into dense element annotations while recording the outcome of each executed action. Using this environment, we construct \textbf{\dataset}, a corpus of 1.2M annotated desktop observations containing 159.7M element instances. We fine-tune four vision-language models on 200K grounding examples drawn from \dataset. All four improve across held-out desktop conditions and on all five external GUI grounding benchmarks; for Qwen3.5-4B, accuracy increases by 11.51 percentage points on ScreenSpot-Pro and 10.11 points on OSWorld-G. The gains also translate to long-horizon task completion: under a fixed planner, the fine-tuned action models solve more WebArena-Infinity and OpenApps tasks, with Qwen3.5-4B increasing from 31 to 50 of 119 tasks and from 3 to 15 of 100 tasks, respectively. These results show that controllable composition of real desktop environments provides a scalable source of supervision for improving both GUI grounding and long-horizon computer use. The framework code, the dataset, and the fine-tuned model are available from the project page: \url{https://saidgurbuz.github.io/deskforge/}.
\end{abstract}

\section{Introduction}
Desktop interfaces are compositional: application windows, dialogs, and system controls coexist on the same screen. Computer-use agents must ground actions within this dense visual context, distinguishing relevant controls from similar elements and cross-application distractors. The same application can present different content or occupy a different position within a multi-window layout. This motivates training resources that capture complete desktop scenes, retain detailed information about their structure, and support systematic variation across desktop configurations.

\begin{figure}[t]
\centering
\includegraphics[width=\linewidth]{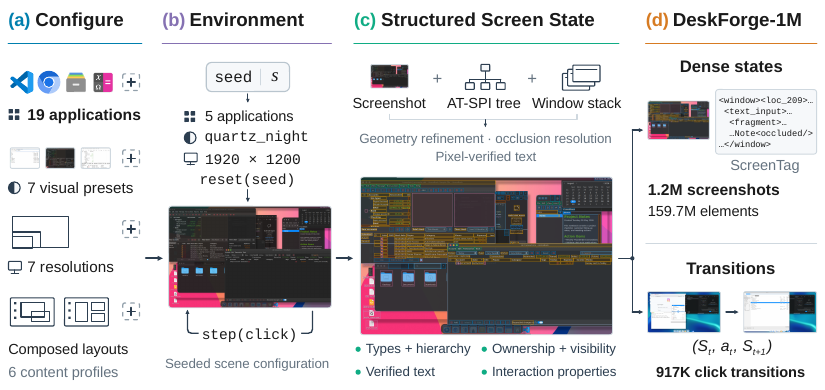}
\caption{\textbf{Overview of \framework\ and \dataset.} (a) A scene specification configures the desktop. (b) Real applications are launched and explored. (c) Screenshots, accessibility trees, and window stacks yield dense annotations. (d) Annotated states and transitions form \dataset.}
\label{fig:pipeline}
\end{figure}

Recent work has expanded GUI supervision through structured screenshot corpora, large-scale grounding datasets, synthetic interfaces, and human demonstrations~\citep{rico,webui,screenparse,uground,osatlas,jedi,groundcua}, while interactive environments make it possible to study agents and tasks under controlled application conditions~\citep{openapps,cuagym,webarena_infinity}. Each resource provides some of these properties, but none combines them: static corpora cannot vary the scenes they capture, most annotate isolated applications or single targets, and executable environments are designed for task evaluation rather than dense screen annotation (Table~\ref{tab:datasets}).

Our key idea is to generate the desktop itself rather than record it: when real applications are composed in controlled scenes, every screen is captured together with its interface structure and window layout, every executed action with its outcome, and the factors that make desktop grounding hard can be varied deliberately. We implement this idea in \framework, a Linux desktop environment (Figure~\ref{fig:pipeline}). Real applications run in isolated sessions, with scene specifications controlling application selection and initial states, content, window layout and stacking, appearance, and resolution. Obtaining this supervision requires more than arranging application windows: accessibility APIs expose application-dependent and often incomplete interface structure, while rendered appearance, window relationships, and action outcomes require separate observations. The annotation pipeline therefore reconciles accessibility information, screenshots, and window geometry to construct dense annotations of element geometry, text, visibility, and interaction properties. It also retains interface hierarchy, application and window ownership, and stacking relationships, placing individual elements within the structure of the composed desktop. Executed clicks are linked to successor observations, recording how the interface changes through interaction.

Using \framework, we construct \dataset, a corpus of 1.2M annotated desktop observations containing 159.7M element instances and 917K recorded click transitions. It spans 19 applications, seven appearance presets, and seven display resolutions. Scene-level splits keep related captures together and test generalization to new scenes, held-out applications, a reserved appearance preset, and a reserved resolution. Natural-language instructions derived from recorded interactions provide a grounding view of the corpus, pairing each before screenshot and instruction with a single target.

Fine-tuning four vision-language models on \dataset\ grounding examples improves accuracy in every held-out condition and raises mean accuracy across five external GUI grounding benchmarks by 1.5 to 24.9 points. For Qwen, mean held-out accuracy rises from 76.26\% to 87.54\%, and the gain grows as more applications share the screen and as targets become partly occluded, the conditions that make desktop grounding hard. Externally, Qwen gains 11.51 percentage points on ScreenSpot-Pro and 10.11 on OSWorld-G. The dense annotations also train a whole-screen element detector: an RT-DETRv4-L model fine-tuned on \dataset\ outperforms the OmniParser v2 detector and ScreenParse YOLO11L on GroundCUA across three localization metrics.

To assess downstream utility, we fix a Qwen3.6-27B planner~\citep{qwen36} and change only the action-model weights within each base/fine-tuned comparison. All four fine-tuned action models solve more tasks on a custom 119-task WebArena-Infinity panel and the 100-task OpenApps longer-horizon set. For Qwen, tasks solved increase from 31 to 50 and from 3 to 15, respectively. These comparisons connect single-target grounding supervision to improved multistep task completion within the tested planner-based agent. Our contributions are:
\begin{enumerate}
\item \textbf{\framework, a controllable desktop environment.} 
\framework\ composes and explores real applications under configurable desktop conditions, collecting dense structured observations and recorded interactions at scale.

\item \textbf{\dataset, a large-scale densely annotated desktop corpus.} 
The corpus contains 1.2M annotated observations, 159.7M element instances, and 917K recorded click transitions, combining dense per-screen annotations with action-linked before/after observations.

\item \textbf{Grounding gains and downstream task-completion benefits.} Fine-tuning four vision-language models improves held-out and external grounding and increases task completion under a fixed planner.
\end{enumerate}

\section{Related Work}
\label{sec:related}

\paragraph{Structured screen supervision.}
GUI supervision ranges from locating individual targets to recovering structured descriptions of a screen. Rico and WebUI pair screenshots with mobile view hierarchies and webpage metadata, respectively~\citep{rico,webui}. Screenshot-to-structure learning builds on this information: Pix2Struct predicts simplified HTML from masked screenshots, while ScreenParse provides dense web-screen annotations and introduces the ScreenTag representation adopted here~\citep{pix2struct,screenparse,smoldocling}. GroundCUA brings dense, human-verified annotation to expert desktop workflows~\citep{groundcua}. These resources motivate retaining element attributes and relationships beyond an action's target, but annotate individual web pages, mobile screens, or applications rather than composed multi-application desktops.\vspace{-5pt}

\paragraph{Automated collection and interface synthesis.}
Automated pipelines expand interface coverage~\citep{seeclick,osatlas,scalecua}, while UGround synthesizes referring expressions from webpage structure and visual content~\citep{uground}. Other pipelines vary the interfaces themselves: Jedi combines UI decomposition with synthesis and augmentation, while MolmoPoint-GUISyn renders generated HTML interfaces~\citep{jedi,molmopoint_guisyn}. GUIrilla explores running applications through native accessibility APIs to build state-action graphs, but processes each application individually~\citep{guirilla}. \framework\ instead configures and explores multiple applications within shared desktop scenes, varying their joint layout and surrounding context.\vspace{-5pt}

\paragraph{Controllable environments and desktop composition.}
Executable environments connect interface actions to task-level outcomes, as in WebArena and OSWorld~\citep{webarena,osworld}. Environment control also supports different forms of variation: OpenApps changes application appearance and content to study reliability~\citep{openapps}, while CUA-Gym and WebArena-Infinity construct environment states or applications together with tasks and verifiers~\citep{cuagym,webarena_infinity}. WinDeskGround composes captured window images to vary layout, occlusion, and distraction for grounding evaluation, but its static compositions cannot execute further actions~\citep{windeskground}. \framework\ performs composition and interaction within a running desktop, collecting dense annotations together with observed click outcomes.\vspace{-5pt}

\paragraph{Interaction data and transition supervision.}
Interaction records connect screen observations to task-directed behavior. Mind2Web and AgentNet provide human demonstrations, while VideoCUA within CUA-Suite preserves continuous expert interaction~\citep{mind2web,opencua,cuasuite}. GUI-360 collects trajectories through automated task execution and retains screenshots and accessibility metadata~\citep{gui360}. An alternative reverses the order of interaction and task specification: OS-Genesis explores first and retrospectively derives tasks~\citep{osgenesis}. \framework\ uses this interaction-first approach to construct single-step grounding instructions from recorded clicks and before/after observations. Recorded transitions can also support joint inverse and forward dynamics learning~\citep{statetransitions}; our corpus retains them for such objectives.

Table~\ref{tab:datasets} summarizes the comparison: among the listed resources, only \framework\ combines dense labels, multi-application scenes, a configurable environment, visibility geometry, and recorded transitions.
\begin{table}[t]
\setlength{\abovecaptionskip}{3pt}
\centering
\small
\caption{\textbf{Scale and capabilities of GUI resources.} The \framework\ row combines its environment capabilities with the dataset counts of \dataset.}
\label{tab:datasets}
\setlength{\tabcolsep}{3pt}
\renewcommand{\arraystretch}{0.95}
\renewcommand{\tabularxcolumn}[1]{m{#1}}
\begin{tabularx}{\linewidth}{@{}lrr*{2}{>{\centering\arraybackslash\hsize=0.95\hsize}X}>{\centering\arraybackslash\hsize=1.2\hsize}X*{2}{>{\centering\arraybackslash\hsize=0.95\hsize}X}@{}}
\toprule
\multicolumn{1}{@{}l}{} & \multicolumn{2}{c}{Scale} & \multicolumn{5}{c}{Capabilities} \\
\cmidrule(lr){2-3}\cmidrule(l){4-8}
Resource & Images & Elements & \shortstack{Dense\\labels} & \shortstack{Multi-app\\scenes} & {\shortstack{Configurable\\Environment}} & \shortstack{Visibility\\geometry} & \shortstack{Recorded\\transitions} \\
\midrule
UGround & 1.3M & 10M & \xmark & \xmark & \xmark & \xmark & \xmark \\
WinDeskGround & 585$^{\dagger}$ & 1,356$^{\ddagger}$ & \xmark & \cmark & \xmark & \cmark & \xmark \\
MolmoPoint-GUISyn & 36K & 2M & \cmark & \xmark & \xmark & \xmark & \xmark \\
AgentNet & 421K & 421K & \xmark & \cmark & \xmark & \xmark & \cmark \\
GroundCUA & 55.6K & 3.56M & \cmark & \xmark & \xmark & \xmark & \xmark \\
ScreenParse & 771K & 21M & \cmark & \xmark & \xmark & \xmark & \xmark \\
OS-Atlas & 2.3M & 13M & \xmark & \cmark & \xmark & \xmark & \xmark \\
ScaleCUA & 1.6M & 17.1M & \cmark & \xmark & \xmark & \xmark & \cmark \\
\midrule
\textbf{\framework} & \textbf{1.21M} & \textbf{159.7M} & \cmark & \cmark & \cmark & \cmark & \cmark \\
\bottomrule
\end{tabularx}
\par\smallskip\begin{minipage}{\linewidth}\footnotesize
\cmark\ supported; \xmark\ not supported; $^{\dagger}$source window images; $^{\ddagger}$target annotations. Definitions: Appendix~\ref{app:comparison}.
\end{minipage}
\end{table}

\section{The \framework\ Environment}
\label{sec:method}

\framework\ combines controllable desktop execution with structured observation to generate dense screen annotations and recorded interactions (Figure~\ref{fig:pipeline}). Applications render their own interfaces within configured desktop scenes; the annotation pipeline then relates the rendered content to interface structure and records the outcomes of executed actions.

\subsection{Scene composition and visual diversity}

Each desktop scene is defined by a specification that selects applications and their initial states, prepares their content, and sets window layout and stacking, appearance, and display resolution. Sampling weights control the prevalence of these settings, while layout constraints define feasible multi-window arrangements. This places the same application in different contexts, with competing controls, labels, and partially overlapping windows. Application content is also configurable, so variation extends beyond window placement to the information presented within the interface. Each axis is extensible: appearance presets and resolutions are configuration entries, and an application is added through a manifest that specifies its launch command, staged content, and scripted initial states, provided its accessibility tree exposes the main interface content (Appendix~\ref{app:generation}).

\begin{figure}[t]
\centering
\setlength{\tabcolsep}{3pt}
\begin{tabular}{@{}cc@{}}
\includegraphics[width=.47\textwidth]{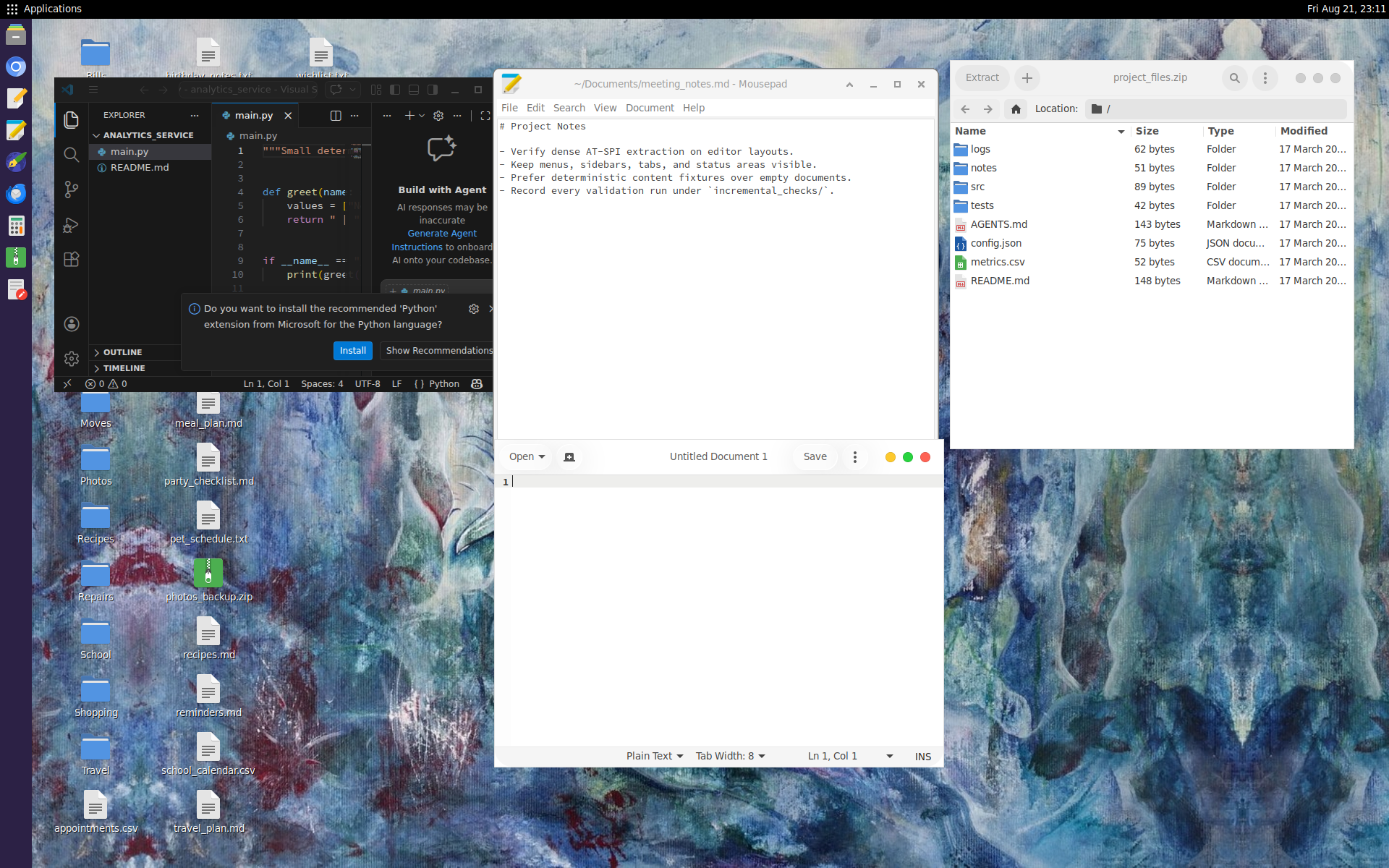} &
\includegraphics[width=.47\textwidth]{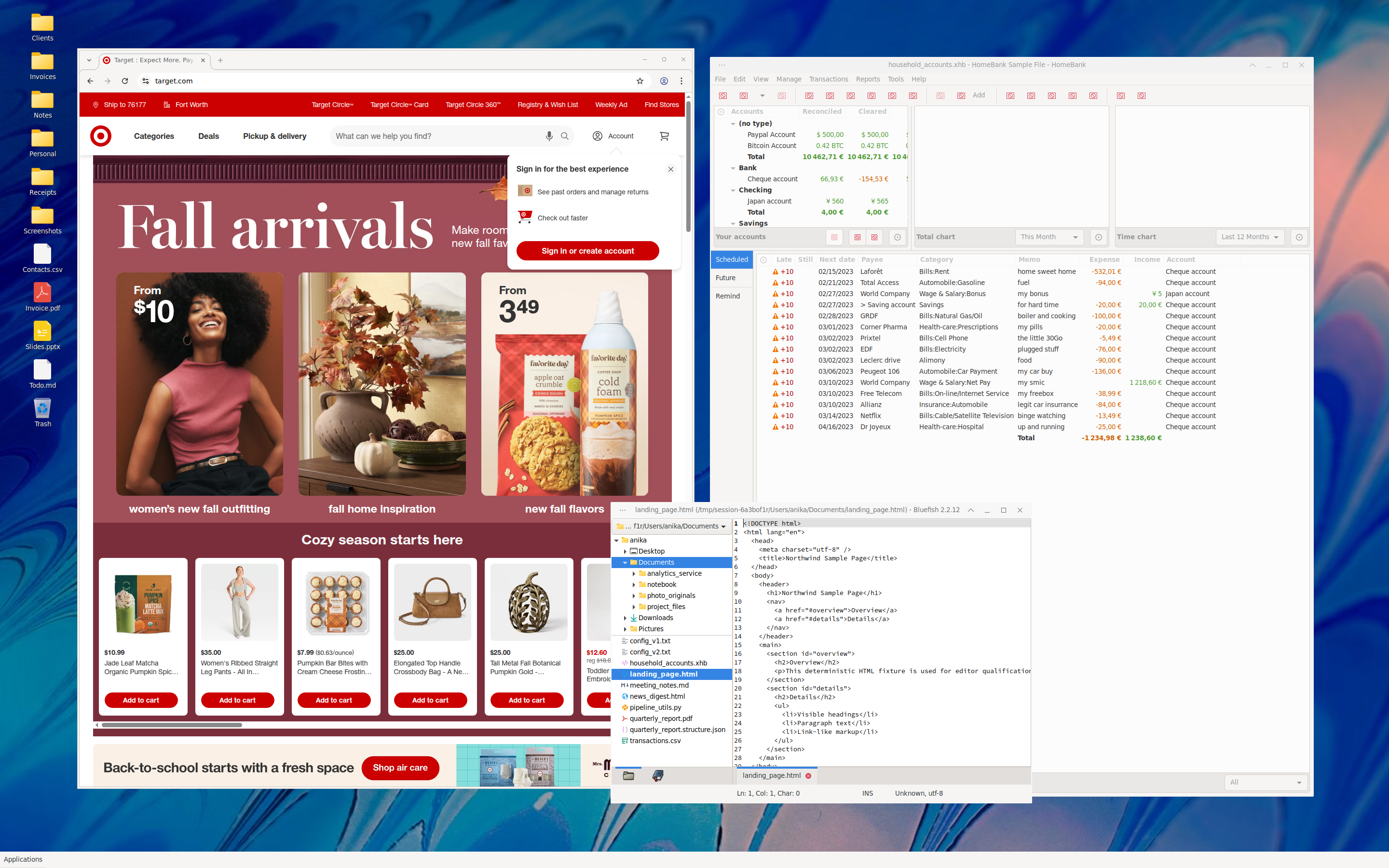}\\
\small Ubuntu-like preset & \small Windows Redmond preset\\[1mm]
\includegraphics[width=.47\textwidth]{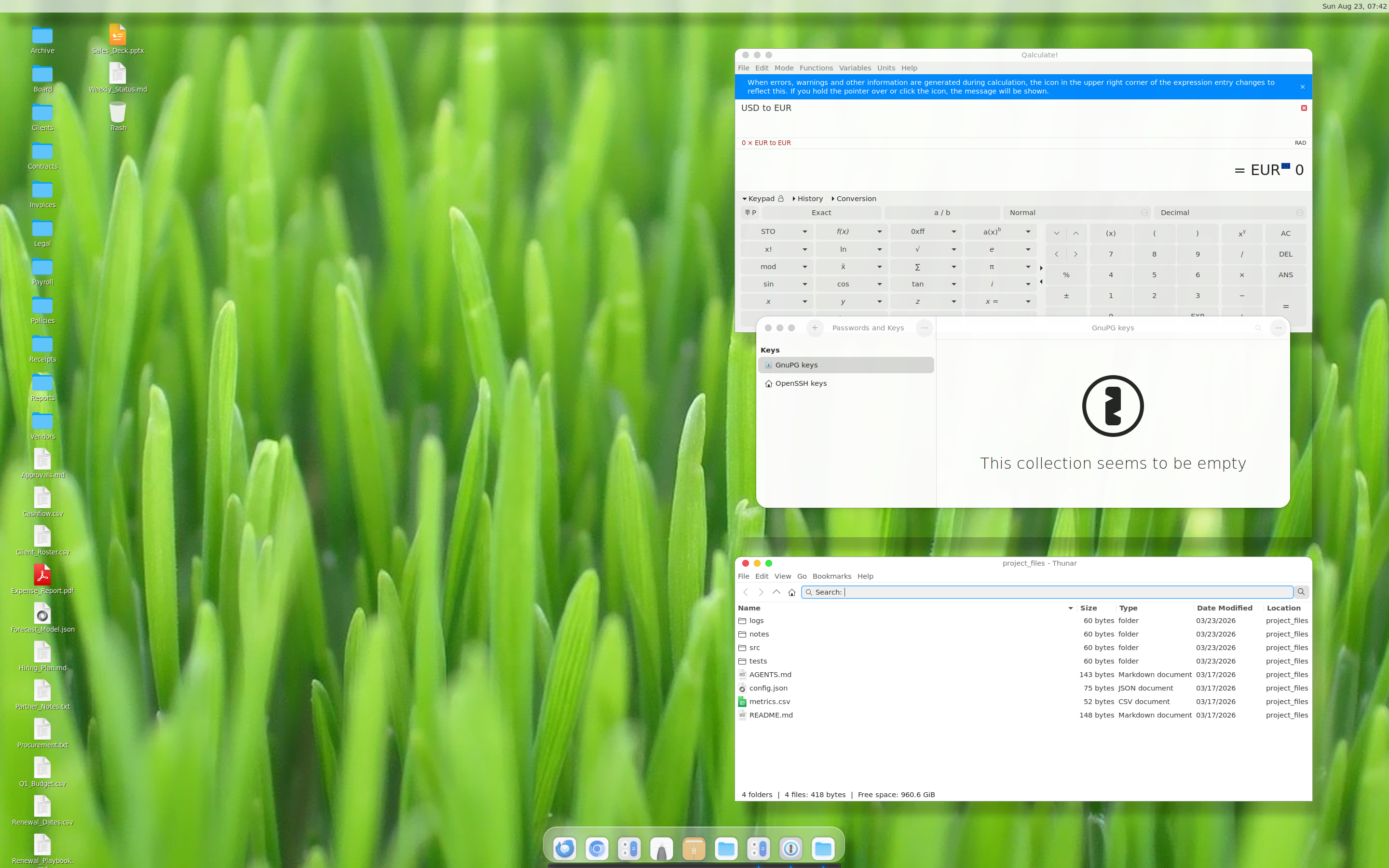} &
\includegraphics[width=.47\textwidth]{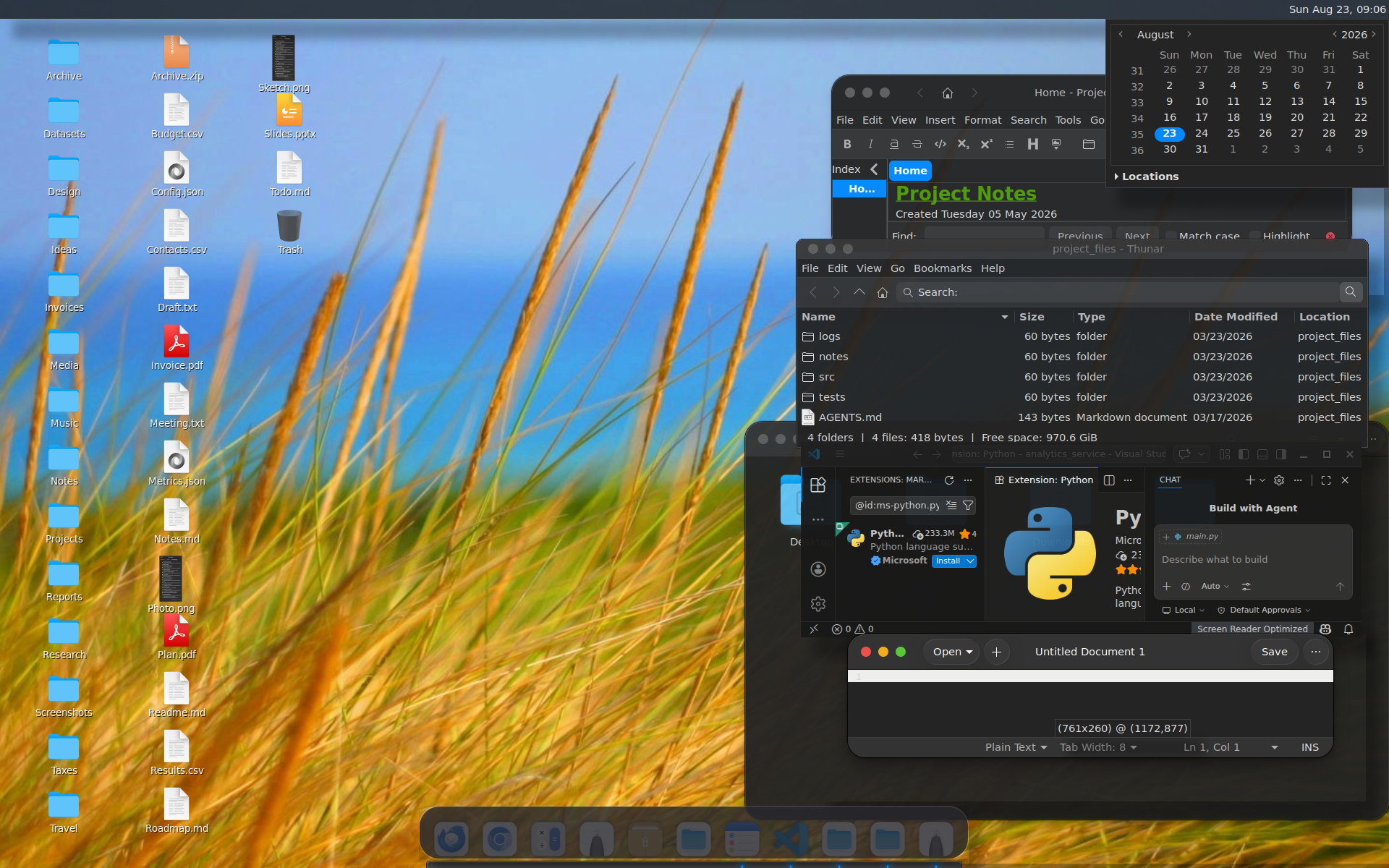}\\
\small macOS Tahoe-like preset & \small Quartz Night preset
\end{tabular}
\vspace{-3pt}
\caption{\textbf{Appearance variation within \framework.} Representative captured desktops illustrate differences in application styling, iconography, window decorations, and panel or dock layout.}
\label{fig:appearance}
\end{figure}

To broaden visual coverage beyond a single Linux desktop style, we integrate community-developed application themes, icon sets, and window decorations with configurable panels, docks, and backgrounds. Seven appearance presets span classic Linux, Ubuntu-like, Windows-inspired, and macOS-inspired styles, including light and dark variants. These assets configure the running desktop and application toolkits, varying widget appearance, iconography, and desktop layout rather than only the background image. Figure~\ref{fig:appearance} illustrates the resulting visual variation. All presets share a Linux execution backend; they reproduce selected visual conventions rather than execute native Windows or macOS applications. Preset definitions and asset sources are given in Appendix~\ref{app:appearance}.

\subsection{Dense structured annotations}

The annotation pipeline combines screenshots, accessibility observations, and measured window relationships. Its output describes element types, text, geometry, and interaction properties together with parent--child structure, reading order, and application and window ownership. Accessibility-reported actions and control states, such as whether an element is enabled, selected, or expanded, supplement the visual annotations with information about the captured interface.

Accessibility structure alone does not determine what is visible. Geometry refinement and clipping account for element boundaries, ancestor containers, viewport limits, overlapping windows, and transient overlays. For partially covered elements, visible support is represented as a union of rectangles, preserving the exposed portions of a control rather than treating its entire extent as a valid click region. Accessibility names are kept separate from displayed text: character geometry and pixel checks assess which text is exposed, so an icon's semantic name is not automatically treated as text rendered on screen. Selected desktop controls omitted from accessibility trees are recovered from window geometry.

A visible, redundancy-reduced annotation view is the basis for screen parsing and ScreenTag serialization~\citep{screenparse}, retaining structural nodes where needed. The environment separately preserves captured hidden elements and their visibility states, allowing the retained interface representation to extend beyond the currently exposed content. Automated audits exclude the 4.7\% of captures with inconsistent or implausibly sparse annotations, and pixel-level checks for missed and spurious annotations guided the pipeline's development. A human audit further confirms annotation quality, finding 99.8\% of sampled element annotations correct. Annotation views, refinement, and quality assessment are detailed in Appendices~\ref{app:release}, \ref{app:implementation}, and~\ref{app:audit}.

\subsection{Interaction recording and instruction construction}

\framework\ explores scenes by sampling actionable elements and executing clicks within their estimated visible regions. Exploration is driven by element selection rather than predefined task goals, and records both changed and unchanged outcomes. Each transition $(S_t,a_t,S_{t+1})$ links before and after screen observations with the executed action, target metadata, and an aggregate effect summary. The associated element records also allow changes in content, geometry, visibility, and control state to be derived separately.

To construct grounding examples, we prompt a vision-language model with the before and after observations and the target context to write a single-step instruction. Instructions are refined and filtered to describe a goal whose target can be identified from the before screenshot, without referring to coordinates or annotation markers. The resulting grounding example pairs that screenshot and instruction with the recorded target. The successor observation informs instruction construction but is not supplied to the grounding model. This yields instruction-grounding supervision while retaining the underlying transitions for other learning objectives.

Acquisition details and instruction-generation prompts are provided in Appendices~\ref{app:generation} and~\ref{app:prompts}.

\section{The \dataset\ Dataset}
\label{sec:dataset}

\subsection{Scale and composition}
\dataset\ contains 1.21M annotated desktop observations and 159.7M element instances, spanning 19 applications, seven appearance presets, and seven display resolutions. The applications include browsers, editors, file managers, and desktop utilities, while resolutions range from $1366\times768$ to $3840\times2160$. Observations are grouped into approximately 324K scenes and include individual captures and interaction episodes, with 917K recorded click transitions. Element counts are taken from the visible annotation view, after removing redundant entries, and count each element once per observation. The total is about 7.6 times that of the largest resource in Table~\ref{tab:datasets}.

\begin{figure}[t]
\centering
\includegraphics[width=\linewidth,trim=0 6pt 0 6pt,clip]{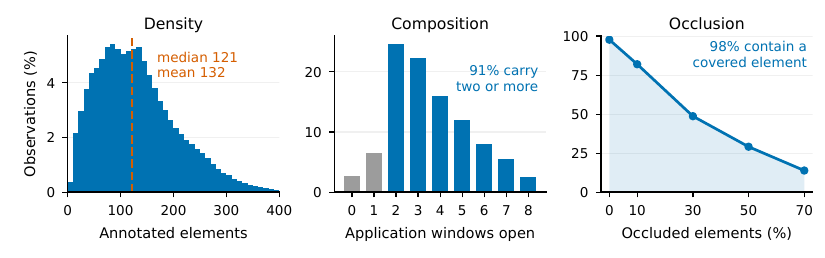}
\vspace{-12pt}
\caption{\textbf{Corpus composition of \dataset.} Over all 1.21M observations: annotated elements per observation (truncated at 400; maximum 1,161), number of applications, and share of observations with more than $x\%$ of elements occluded.}
\label{fig:composition}
\end{figure}

As Figure~\ref{fig:composition} shows, observations are dense and cluttered: they contain 132 annotated elements on average (median 121), 90.9\% record at least two applications, and 97.7\% contain occluded elements, with more than 30\% of elements affected in nearly half. These statistics describe the desktop context rather than the visibility of the targets selected for grounding.

\subsection{Annotation content and supervision views}
Each observation pairs a screenshot with dense element records and a ScreenTag representation. The element records give each element's geometry, text, type, window ownership, and interaction properties, and store both its full extent and its visible region, the fragments left exposed by overlapping windows; ScreenTag serializes the same screen as compact markup that preserves the element hierarchy (Section~\ref{sec:method}). Transition records add the executed click, its target, the successor observation, and a summary of what changed.

To introduce linguistic variation into the grounding supervision, we use Qwen3.6-27B to synthesize natural-language instructions from programmatically recorded clicks, target context, and before/after screenshots. The model is prompted to express the same single-step user goal in concise, standard, and more contextual formulations, varying wording and detail while preserving the intended target. Full prompts are provided in Appendix~\ref{app:prompts}.

The grounding experiments use a fixed 200K-example training view, with each example pairing a before screenshot and one synthesized instruction with the recorded click target. Element detection (Section~\ref{sec:detection}) instead uses the dense annotations of its training screens.

In a human audit, 97.8\% of uniformly sampled instructions were judged sound (Appendix~\ref{app:audit}).

\subsection{Evaluation splits}
Training, validation, and test partitions are assigned at the scene level, keeping interaction frames and recaptures of the same scene together. Four test conditions distinguish new scenes with represented desktop attributes (New Scenes) from scenes containing held-out applications (App), a reserved appearance preset (Theme), or a reserved display resolution (Resolution).

The application split holds out GNOME System Monitor, Pluma, and Xarchiver. Assignment considers every application recorded across a scene, rather than only the application containing a selected target, so a held-out application cannot enter training through another frame. The appearance split reserves the Quartz Night Nord preset, and the resolution split reserves $2880\times1800$. These partitions evaluate generalization to desktop configurations not represented in training. The full application inventory, corpus distributions, and split specifications are provided in Appendix~\ref{app:release}.

\section{Experiments}
\label{sec:experiments}
We fine-tune vision-language models on \dataset\ and ask three questions: does grounding improve on held-out desktop configurations, do the gains transfer to external GUI benchmarks, and does long-horizon task completion increase when only the action model changes within a fixed planner agent? We then test whether the dense annotations also support whole-screen element detection. These comparisons assess the utility of the generated data beyond the scenes and individual interactions used for fine-tuning.

\begin{table}[t]
\centering
\small
\caption{\textbf{Grounding accuracy on the four held-out conditions of \dataset\ (\%).}
Public checkpoints provide reference performance. The lower block directly compares each backbone before and after fine-tuning on \dataset; fine-tuned results are shown in bold.}
\label{tab:publicpanel}

\setlength{\tabcolsep}{4.5pt}
\renewcommand{\arraystretch}{1.06}

\sisetup{
  table-number-alignment = center,
  table-format = 2.2,
  detect-weight = true,
  detect-family = true
}

\begin{tabularx}{\linewidth}{
  @{}
  >{\raggedright\arraybackslash}p{1.35in}
  >{\raggedright\arraybackslash}X
  *{5}{S}
  @{}
}
\toprule

\multirow{2}{*}{\textbf{Model}}
& \multirow{2}{*}{\textbf{Variant}}
& \multicolumn{4}{c}{\textbf{Held-out condition}}
& \multicolumn{1}{c}{\multirow{2}{*}{\textbf{Mean}}} \\

\cmidrule(lr){3-6}

&
& {\textbf{New Scenes}}
& {\textbf{Theme}}
& {\textbf{App}}
& {\textbf{Resolution}}
& {} \\

\midrule

\multicolumn{7}{@{}l}{\itshape Public VLMs and CUAs} \\
\addlinespace[1pt]

OS-Atlas-Pro-7B
& & 47.61 & 47.27 & 45.58 & 32.10 & 43.14 \\

EvoCUA-8B
& & 50.49 & 45.59 & 46.65 & 44.93 & 46.91 \\

UI-TARS-1.5-7B
& & 67.68 & 64.96 & 66.68 & 61.67 & 65.25 \\

GUI-Owl-1.5-32B
& & 70.87 & 62.19 & 66.68 & 64.65 & 66.10 \\

UGround-V1-7B
& & 69.08 & 64.61 & 67.18 & 63.82 & 66.17 \\

ScaleCUA-7B
& & 72.74 & 68.28 & 70.75 & 63.62 & 68.85 \\

UI-MOPD-8B
& & 73.04 & 67.12 & 70.49 & 68.07 & 69.68 \\

GroundNext-7B
& & 73.57 & 67.67 & 72.10 & 68.87 & 70.55 \\

Gemma4-31B
& & 75.26 & 71.00 & 73.57 & 63.13 & 70.74 \\

UI-Venus-2-9B
& & 81.45 & 77.44 & 79.40 & 77.40 & 78.92 \\

\midrule

\multicolumn{7}{@{}l}{\itshape Fine-tuning across VLMs} \\
\addlinespace[1pt]

\multirow{2}{*}{\textbf{Gemma4-E4B}}
& Base
& 16.27 & 12.10 & 14.15 & 6.56 & 12.27 \\
& \textbf{+ \dataset}
& \bfseries 89.01
& \bfseries 85.36
& \bfseries 84.50
& \bfseries 81.62
& \bfseries 85.12 \\

\midrule[0.1pt]
\multirow{2}{*}{\textbf{UI-R1-3B}}
& Base
& 41.34 & 40.24 & 35.56 & 34.90 & 38.01 \\
& \textbf{+ \dataset}
& \bfseries 88.25
& \bfseries 84.28
& \bfseries 83.77
& \bfseries 82.07
& \bfseries 84.59 \\

\midrule[0.1pt]
\multirow{2}{*}{\textbf{InternVL3.5-8B}}
& Base
& 65.00 & 59.28 & 61.50 & 47.62 & 58.35 \\
& \textbf{+ \dataset}
& \bfseries 87.37
& \bfseries 83.43
& \bfseries 83.77
& \bfseries 71.10
& \bfseries 81.42 \\

\midrule[0.1pt]
\multirow{2}{*}{\textbf{Qwen3.5-4B}}
& Base
& 78.56 & 74.03 & 76.11 & 76.34 & 76.26 \\
& \textbf{+ \dataset}
& \bfseries 90.14
& \bfseries 87.36
& \bfseries 87.50
& \bfseries 85.15
& \bfseries 87.54 \\

\bottomrule
\end{tabularx}
\end{table}

\paragraph{Experimental setup.}
We fine-tune Qwen3.5-4B~\citep{qwen35}, Gemma4-E4B IT~\citep{gemma4}, InternVL3.5-8B~\citep{internvl35}, and UI-R1-3B~\citep{uir1} on the 200K single-target grounding view of \dataset\ using LoRA~\citep{lora}. The first three are general-purpose VLMs; UI-R1 provides an additional test starting from a GUI-specialized checkpoint. Within each base/fine-tuned pair, prompts, image preprocessing, decoding, and scoring are held fixed, while model-specific input and output conventions are retained.

Grounding accuracy measures whether a predicted point falls within the annotated target region, with invalid outputs counted as incorrect. DeskForge uses the annotated visible regions; external evaluation uses the corresponding benchmark annotations and scoring settings. Unless otherwise specified, fine-tuning uses eight NVIDIA H100 GPUs for one epoch, with an effective global batch size of 64. Full training settings and model-specific exceptions are provided in Appendix~\ref{app:recipes}.

\subsection{Grounding across held-out desktop conditions}
We first evaluate the four scene-level conditions defined in Section~\ref{sec:dataset}: New Scenes, Theme, App, and Resolution. Table~\ref{tab:publicpanel} reports each condition separately and the mean over the four conditions. The matched base/fine-tuned pairs measure the benefit of fine-tuning, while public checkpoints~\citep{osatlas,evocua,uitars,guiowl,uground,scalecua,uimopd,groundcua,gemma4,uivenus2} provide reference performance under model-specific, rather than equal-compute, inference protocols (Appendix~\ref{app:publicprotocol}).

All four models improve in every held-out condition (Table~\ref{tab:publicpanel}). Qwen's mean accuracy rises from 76.26\% to 87.54\%, while UI-R1 improves from 38.01\% to 84.59\%. The fine-tuned Gemma and InternVL models reach 85.12\% and 81.42\%, respectively.

The gains hold for held-out applications and the reserved appearance and resolution, so the supervision transfers beyond the desktop configurations seen in training. They are also largest where desktop scenes are hardest (Appendices~\ref{app:robustness} and~\ref{app:grounding-failures}). As the number of applications on screen grows from one to four or more, fine-tuned Qwen loses only a few points, whereas its base model and UI-Venus-2-9B, the strongest reference model, degrade steadily; the gap to the base widens from 9.7 to 14.3 points. The gap likewise grows from 10.9 to 13.7 points as the target's visible-area loss rises to 35\%.

\begin{table}[t]
\centering
\small
\caption{\textbf{GUI grounding accuracy on five external benchmarks (\%).}
Rows marked \textbf{+ \dataset} denote fine-tuning on our data.}
\label{tab:external}
\setlength{\tabcolsep}{4.5pt}
\renewcommand{\arraystretch}{1.08}
\sisetup{
  table-number-alignment = center,
  table-format = 2.2,
  detect-weight = true,
  detect-family = true
}
\begin{tabularx}{\linewidth}{@{}>{\raggedright\arraybackslash}p{1.00in} >{\raggedright\arraybackslash}X *{5}{S}@{}}
\toprule
\textbf{Model} & \textbf{Training} & {\textbf{SS-Pro}} & {\textbf{SS-v2}} & {\textbf{OSWorld-G}} & {\textbf{UI-Vision}} & {\textbf{MMB-GUI}} \\
\midrule
\multirow{2}{*}{\textbf{Gemma4-E4B}}
  & Base & 1.83 & 46.86 & 10.64 & 3.28 & 31.39 \\
  & \textbf{+ \dataset} & \bfseries 21.95 & \bfseries 79.56 & \bfseries 44.15 & \bfseries 13.87 & \bfseries 58.90 \\
\addlinespace[2pt]
\midrule
\multirow{2}{*}{\textbf{UI-R1-3B}}
  & Base & 14.48 & 81.92 & 34.04 & 12.09 & 56.84 \\
  & \textbf{+ \dataset} & \bfseries 27.20 & \bfseries 83.96 & \bfseries 43.97 & \bfseries 15.15 & \bfseries 65.41 \\
\addlinespace[2pt]
\midrule
\multirow{2}{*}{\textbf{InternVL3.5-8B}}
  & Base & 18.72 & 83.33 & 44.15 & 17.50 & 66.58 \\
  & \textbf{+ \dataset} & \bfseries 23.78 & \bfseries 84.98 & \bfseries 44.33 & \bfseries 17.55 & \bfseries 67.20 \\
\addlinespace[2pt]
\midrule
\multirow{2}{*}{\textbf{Qwen3.5-4B}}
  & Base & 30.17 & 80.11 & 51.24 & 17.24 & 61.83 \\
  & \textbf{+ \dataset} & \bfseries 41.68 & \bfseries 90.33 & \bfseries 61.35 & \bfseries 28.77 & \bfseries 76.82 \\
\bottomrule
\end{tabularx}
\end{table}

\subsection{Transfer to external grounding benchmarks}
To test whether the benefit extends beyond \dataset, we evaluate on ScreenSpot-Pro, ScreenSpot-v2, OSWorld-G, UI-Vision, and MMBench-GUI~\citep{screenspotpro,osatlas,jedi,uivision,mmbenchgui}. None is used for fine-tuning, and all differ from \dataset\ in platform coverage, application inventory, and annotation source: they are human-annotated captures of real usage, spanning Windows, macOS, Linux, mobile, and web screens, and include professional applications such as CAD and creative tools that \framework\ does not run.

All four models obtain higher accuracy on all five external benchmarks (Table~\ref{tab:external}). Qwen gains 11.51 percentage points on ScreenSpot-Pro and 10.11 points on OSWorld-G, while Gemma's mean over the five benchmarks increases by 24.9 points. UI-R1 improves from 14.48\% to 27.20\% on ScreenSpot-Pro and from 39.87\% to 47.14\% in mean accuracy over the five benchmarks. InternVL also improves across all five benchmarks, with a more modest mean gain of 1.51 points. The ScreenSpot-v2 breakdown (Appendix~\ref{app:ssv2}, Table~\ref{tab:ssv2_domains}) shows where transfer occurs: Qwen and UI-R1 gain most on desktop interfaces (+14.37 and +5.69 points), yet also improve on mobile (+10.77 and +0.60) and web (+6.41 and +0.91), although \dataset\ contains no mobile screens.

These results show that supervision generated through controlled desktop execution transfers to GUI collections beyond the DeskForge environment. The gains for UI-R1 are particularly informative: \dataset\ benefits not only general-purpose VLMs, but also a model already specialized for GUI interaction. This demonstrates its value both for developing grounding capabilities and for further improving an existing computer-use model.

\subsection{Long-horizon task completion under a fixed planner}
We next test whether single-target grounding supervision improves task completion during extended interaction. A fixed Qwen3.6-27B planner~\citep{qwen36} selects action types, text or keys, and coordinate-free target descriptions; the action model supplies the locations needed to execute those decisions. Only the action-model weights change within each base/fine-tuned comparison. We evaluate a custom 119-task
WebArena-Infinity panel and the 100-task OpenApps longer-horizon set~\citep{webarena_infinity,openapps}. Task success is determined by the environments' state-based checks; task selection, action budgets, and agent protocols are specified in Appendix~\ref{app:interaction}.

\begin{table}[t]
\centering
\small
\caption{\textbf{Long-horizon task completion with a fixed Qwen3.6-27B planner.}
Number of solved tasks on the custom 119-task WebArena-Infinity panel and
the 100-task OpenApps longer-horizon set. \textbf{+\dataset} denotes
fine-tuning of the model on our data.}
\label{tab:interaction}

\setlength{\tabcolsep}{7pt}
\renewcommand{\arraystretch}{1.08}

\sisetup{
  table-number-alignment = center,
  table-format = 2.0,
  detect-weight = true,
  detect-family = true
}

\begin{tabular}{
  @{}
  l
  S[table-format=2.0]
  S[table-format=2.0]
  @{\hspace{1.8em}}
  S[table-format=1.0]
  S[table-format=2.0]
  @{}
}
\toprule

\multirow{2}{*}{\textbf{Model}}
& \multicolumn{2}{c}{\textbf{WebArena-Infinity (119)}}
& \multicolumn{2}{c}{\textbf{OpenApps (100)}} \\

\cmidrule(lr){2-3}
\cmidrule(l){4-5}

& {\textbf{Base}}
& {\textbf{+\dataset}}
& {\textbf{Base}}
& {\textbf{+\dataset}} \\

\midrule

Gemma4-E4B
& 5  & \bfseries 40
& 2  & \bfseries 12 \\

UI-R1-3B
& 30 & \bfseries 37
& 7  & \bfseries 12 \\

InternVL3.5-8B
& 37 & \bfseries 44
& 4  & \bfseries 11 \\

Qwen3.5-4B
& 31 & \bfseries 50
& 3  & \bfseries 15 \\

\bottomrule
\end{tabular}
\end{table}

All four fine-tuned action models solve more tasks in both environments (Table~\ref{tab:interaction}). Qwen improves from 31 to 50 tasks on WebArena-Infinity and from 3 to 15 on OpenApps. Gemma shows the largest increase on the Infinity panel, from 5 to 40 tasks. InternVL solves seven additional tasks in each environment, while the GUI-specialized UI-R1 solves seven additional Infinity tasks and five additional OpenApps tasks.

Although fine-tuning uses single grounding examples rather than task trajectories, the fine-tuned action models execute the fixed planner's decisions more successfully over successive observations (Appendix~\ref{app:trajectories}), even in browser environments unlike our composed desktops. Improving grounding alone thus raises task completion without planner training, pointing to a practical use of \dataset: strengthening the visual action-execution component of a computer-use agent.

\subsection{Dense element detection}
\label{sec:detection}
The experiments above use one annotated target per screenshot. Dense annotations additionally enable learning to parse the entire screen, localizing every visible element rather than a single instructed target. We fine-tune an RT-DETRv4-L detector on these annotations and evaluate cross-dataset transfer on GroundCUA. Our detector outperforms the OmniParser v2 detector and ScreenParse YOLO11L on all three localization metrics, reaching 62.17\% group F1 compared with 58.55\% and 57.59\%, respectively. These results show that detectors trained on dense screen annotations from \framework\ localize elements on screens beyond \framework, a core step of screen parsing that complements instruction-conditioned grounding. The comparison and metric definitions are provided in Appendix~\ref{app:detection}.

\section{Conclusion and Future Work}
\label{sec:conclusion}
\framework\ makes the composition and exploration of running desktop applications a controllable source of structured supervision. The resulting \dataset\ contains 1.2M annotated observations and 159.7M element instances, together with recorded click transitions. Fine-tuning improves grounding across held-out desktop conditions and five external benchmarks for both general-purpose and GUI-specialized models, and increases long-horizon task completion under a fixed planner. Dense element detection demonstrates a complementary use of the same corpus. These results indicate the value of controllably generated desktop
data for learning and applying visual grounding.

\paragraph{Limitations and future directions.}
Current acquisition uses a Linux backend and relies on applications with suitable accessibility support. Broader application integration, native-platform backends, and complementary visual annotation would extend coverage to interfaces that expose less structure. Our VLM training focuses on single-target grounding, leaving joint learning from interface structure and recorded state changes as a natural next step. In particular, the retained transitions could support forward and inverse dynamics objectives for learning about action outcomes. Developing goal-conditioned multi-application tasks with state-based verification would further extend \framework\ toward online training of long-horizon computer-use agents.

\par\label{paper:main-text-end}
\clearpage
\bibliographystyle{iclr2027_conference}
\bibliography{references}
\clearpage
\appendix

\section{Dataset Specification}
\label{app:release}

\subsection{Observations, annotations, and supervision views}
\dataset\ contains 1,207,368 annotated observations and 159,723,780
retained element instances. The element total counts entries in the visible,
redundancy-reduced annotation view separately for each observation, including
recurring elements across interaction frames. The mean and median are
132.29 and 121 elements per observation.

The observation payload contains the screenshot,
element annotations, ScreenTag serialization, and capture record;
Table~\ref{tab:fields} lists the information stored for each element,
observation, and transition, and Table~\ref{tab:kinds} the element kinds.
The environment additionally retains broader extraction and amodal records;
these are separate from the main observation payload.

Because every screenshot, annotation, and transition is stored, \dataset\
is used without executing the environment: training and evaluation require
no virtual machines, application installation, or environment resets, and
results on the held-out splits are reproducible from the stored files alone.
When a study needs states beyond those recorded, \framework\ can generate them.

\begin{table}[t]
\centering
\small
\caption{\textbf{Information stored for each element, observation, and transition.}
Element records and ScreenTag accompany every screenshot; transition fields
link consecutive observations of an interaction episode.}
\label{tab:fields}
\setlength{\tabcolsep}{4pt}
\renewcommand{\arraystretch}{1.08}
\begin{tabularx}{\linewidth}{@{}lX@{}}
\toprule
Level & Stored information \\
\midrule
Element & Type (55 ScreenTag classes) and kind (26 categories, Table~\ref{tab:kinds}); accessibility role; accessible name, and visible text with its verification status; visible fragments and, where retained, the source extent; occlusion state (none, partial, hidden); owning application and window, stacking index, parent and child links, and reading order; interaction properties (actionable, enabled, focusable, editable, checkable and checked, expandable and expanded, selected) and a click point inside the visible region; an identifier stable within an episode. \\
Observation & Screenshot and ScreenTag serialization; resolution, appearance preset, panel variant, desktop profile, window layout, and scene seed; launched applications; window stack with titles, extents, and stacking order; element and occlusion counts; near-duplicate and no-op flags; split and generator provenance. \\
Transition & Before and after observations; click point; target identifier, role, name, text, and box in the before frame; element-level effects (appeared, disappeared, moved, text or state changed, newly occluded, revealed) with an aggregate change magnitude; training eligibility and exclusion reasons. \\
\bottomrule
\end{tabularx}
\end{table}

\begin{table}[t]
\centering
\small
\caption{\textbf{Element kinds.} Each element also retains its accessibility
role and one of 55 ScreenTag classes. Kinds merge classes that are not visually
distinguishable; for example, check boxes, radio buttons, switches, and toggle
buttons form \emph{toggle}, whose state is stored separately.}
\label{tab:kinds}
\setlength{\tabcolsep}{4pt}
\renewcommand{\arraystretch}{1.08}
\begin{tabularx}{\linewidth}{@{}lX@{}}
\toprule
Group & Kinds \\
\midrule
Structure & window, titlebar, menubar, toolbar, statusbar, sidebar \\
Navigation & menu, menuitem, tab, link \\
Controls & button, toggle, textinput, select, slider, scrollbar, progress \\
Content & text, heading, image, icon, canvas \\
Collections & list, listitem, table, cell \\
\bottomrule
\end{tabularx}
\end{table}

\subsection{Scene-level splits}
All observations and recaptures belonging to a scene share one split.
Application assignment uses the union of applications recorded throughout
the scene. The reserved applications are GNOME System Monitor, Pluma,
and Xarchiver; the reserved appearance is Quartz Night Nord; and the
reserved resolution is $2880\times1800$. Scenes matching multiple holdouts
are assigned in application, appearance, then resolution priority.
The New Scenes condition, stored as \code{test_id}, holds out scene
instances with represented application, appearance, and resolution attributes.
Table~\ref{tab:releasesplits} reports observations, element instances, click
transitions, and grounding examples in each partition. Excluding observations
marked as near-duplicates or no-op frames leaves a filtered state view of
1,067,799 observations and 142,537,053 element instances, from which the
human audit samples (Appendix~\ref{app:audit}).

\begin{table}[t]
\centering
\small
\caption{\textbf{\dataset\ by split.} Element instances refer to the visible,
redundancy-reduced view. Each grounding example pairs one instruction with its
recorded target; the 200,000 training examples are sampled from 551,651
training-split instructions.}
\label{tab:releasesplits}
\setlength{\tabcolsep}{4pt}
\renewcommand{\arraystretch}{1.08}
\begin{tabularx}{\linewidth}{@{}Xrrrr@{}}
\toprule
Split & Observations & Element instances & Click transitions & Grounding examples \\
\midrule
Train & 999,494 & 126,746,711 & 760,850 & 200,000 \\
Validation & 11,279 & 1,448,957 & 8,667 & 6,281 \\
New Scenes & 22,550 & 2,826,805 & 17,383 & 12,589 \\
App & 43,206 & 8,151,157 & 32,848 & 22,729 \\
Theme & 51,627 & 7,991,478 & 38,671 & 27,175 \\
Resolution & 79,212 & 12,558,672 & 58,792 & 43,210 \\
\midrule
\textbf{Total} & \textbf{1,207,368} & \textbf{159,723,780} & \textbf{917,211} & \textbf{311,984} \\
\bottomrule
\end{tabularx}
\end{table}

\subsection{Application and display composition}
\label{app:applications}
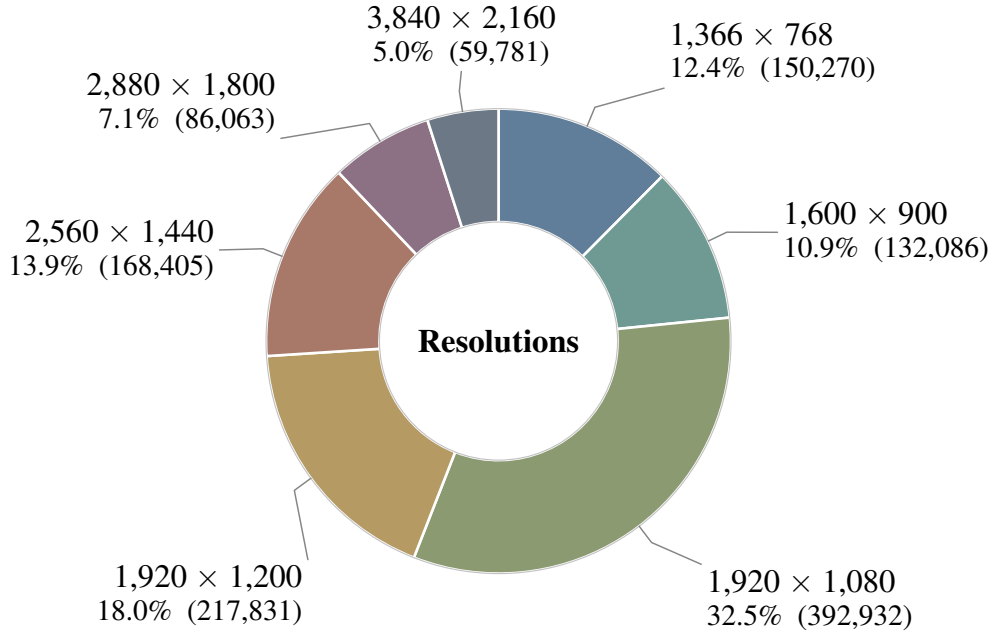
\begin{figure}[t]
\centering
\resizebox{\linewidth}{!}{\definecolor{resA}{HTML}{607D9A} %
\definecolor{resB}{HTML}{6F9A93} %
\definecolor{resC}{HTML}{8B9A71} %
\definecolor{resD}{HTML}{B59A63} %
\definecolor{resE}{HTML}{A87868} %
\definecolor{resF}{HTML}{8C7184} %
\definecolor{resG}{HTML}{6C7885} %

\begin{tikzpicture}[line cap=round,line join=round]
  \def\R{2.50}
  \def\r{1.28}

  \filldraw[fill=resA,draw=white,line width=0.8pt]
    (0,0) -- (90:\R) arc[start angle=90,end angle=45.194108,radius=\R]
    -- (45.194108:\r) arc[start angle=45.194108,end angle=90,radius=\r] -- cycle;
  \filldraw[fill=resB,draw=white,line width=0.8pt]
    (0,0) -- (45.194108:\R) arc[start angle=45.194108,end angle=5.810126,radius=\R]
    -- (5.810126:\r) arc[start angle=5.810126,end angle=45.194108,radius=\r] -- cycle;
  \filldraw[fill=resC,draw=white,line width=0.8pt]
    (0,0) -- (5.810126:\R) arc[start angle=5.810126,end angle=-111.350110,radius=\R]
    -- (-111.350110:\r) arc[start angle=-111.350110,end angle=5.810126,radius=\r] -- cycle;
  \filldraw[fill=resD,draw=white,line width=0.8pt]
    (0,0) -- (-111.350110:\R) arc[start angle=-111.350110,end angle=-176.300614,radius=\R]
    -- (-176.300614:\r) arc[start angle=-176.300614,end angle=-111.350110,radius=\r] -- cycle;
  \filldraw[fill=resE,draw=white,line width=0.8pt]
    (0,0) -- (-176.300614:\R) arc[start angle=-176.300614,end angle=-226.513805,radius=\R]
    -- (-226.513805:\r) arc[start angle=-226.513805,end angle=-176.300614,radius=\r] -- cycle;
  \filldraw[fill=resF,draw=white,line width=0.8pt]
    (0,0) -- (-226.513805:\R) arc[start angle=-226.513805,end angle=-252.175145,radius=\R]
    -- (-252.175145:\r) arc[start angle=-252.175145,end angle=-226.513805,radius=\r] -- cycle;
  \filldraw[fill=resG,draw=white,line width=0.8pt]
    (0,0) -- (-252.175145:\R) arc[start angle=-252.175145,end angle=-270,radius=\R]
    -- (-270:\r) arc[start angle=-270,end angle=-252.175145,radius=\r] -- cycle;

  \draw[black!24,line width=0.35pt] (0,0) circle[radius=\R];
  \draw[black!24,line width=0.35pt] (0,0) circle[radius=\r];

  \node[align=center] at (0,0) {\bfseries Resolutions};

  \draw[black!48,line width=0.45pt] (67.597054:\R) -- (67.597054:2.72) -- (1.78,2.72);
  \node[anchor=south west,align=left,inner sep=1.5pt] at (1.78,2.72)
    {1,366 $\times$ 768\\[-1pt]{\small 12.4\%\enspace (150,270)}};

  \draw[black!48,line width=0.45pt] (25.502117:\R) -- (25.502117:2.72) -- (3.00,1.17);
  \node[anchor=west,align=left,inner sep=1.5pt] at (3.00,1.17)
    {1,600 $\times$ 900\\[-1pt]{\small 10.9\%\enspace (132,086)}};

  \draw[black!48,line width=0.45pt] (-52.769992:\R) -- (-52.769992:2.72) -- (2.18,-2.40);
  \node[anchor=north west,align=left,inner sep=1.5pt] at (2.18,-2.40)
    {1,920 $\times$ 1,080\\[-1pt]{\small 32.5\%\enspace (392,932)}};

  \draw[black!48,line width=0.45pt] (-143.825362:\R) -- (-143.825362:2.72) -- (-2.05,-2.35);
  \node[anchor=north east,align=right,inner sep=1.5pt] at (-2.05,-2.35)
    {1,920 $\times$ 1,200\\[-1pt]{\small 18.0\%\enspace (217,831)}};

  \draw[black!48,line width=0.45pt] (158.592790:\R) -- (158.592790:2.72) -- (-3.00,0.98);
  \node[anchor=east,align=right,inner sep=1.5pt] at (-3.00,0.98)
    {2,560 $\times$ 1,440\\[-1pt]{\small 13.9\%\enspace (168,405)}};

  \draw[black!48,line width=0.45pt] (120.655525:\R) -- (120.655525:2.72) -- (-2.34,2.18);
  \node[anchor=south east,align=right,inner sep=1.5pt] at (-2.34,2.18)
    {2,880 $\times$ 1,800\\[-1pt]{\small 7.1\%\enspace (86,063)}};

  \draw[black!48,line width=0.45pt] (98.912428:\R) -- (98.912428:2.72) -- (-0.40,2.90);
  \node[anchor=south,align=center,inner sep=1.5pt] at (-0.40,2.90)
    {3,840 $\times$ 2,160\\[-1pt]{\small 5.0\%\enspace (59,781)}};
\end{tikzpicture}}
\caption{\textbf{Display resolutions in \dataset.}
Observation counts and proportions across the seven resolutions. All
1,207,368 observations are included, including successive episode frames.}
\label{fig:resolutions}
\end{figure}

\begin{figure}[p]
\centering
\resizebox{\linewidth}{!}{\input{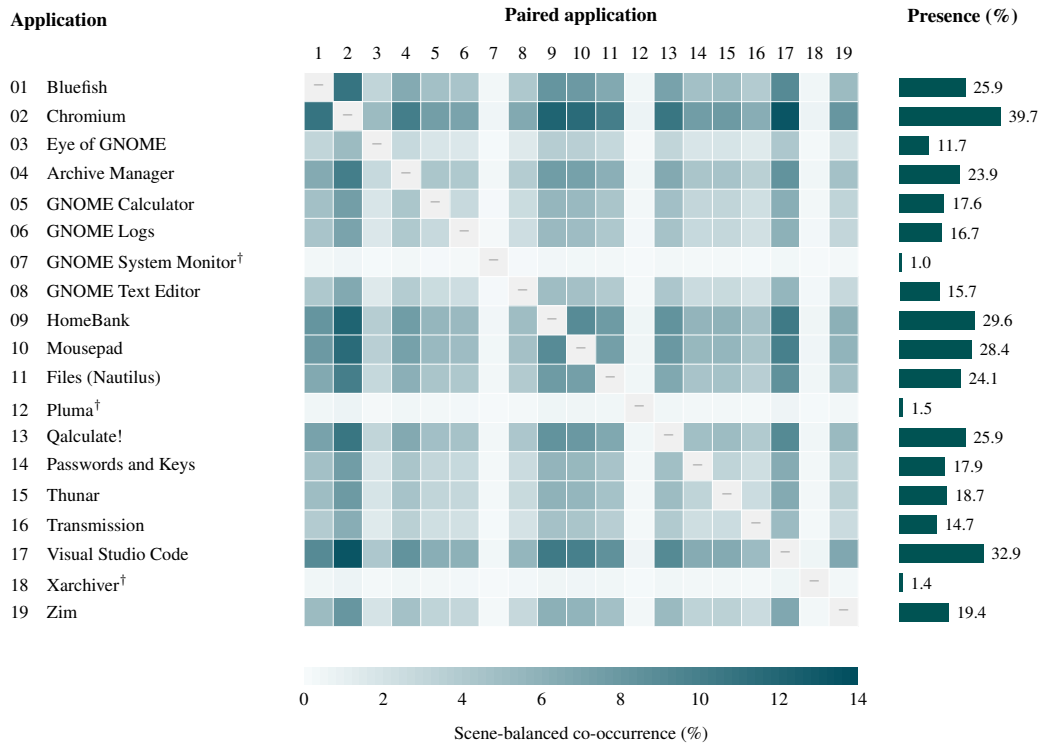}}
\caption{\textbf{Application inventory and co-occurrence.}
Each matrix entry averages the within-scene fraction of captures recording
both applications; bars show the corresponding marginal application presence.
The analysis covers 1,175,179 captures from 315,705 scenes with nonempty
application records. Application presence is taken from capture metadata,
not from visible-window counts. $^{\dagger}$Reserved applications.}
\label{fig:applications}
\end{figure}

Figures~\ref{fig:applications} and~\ref{fig:resolutions} show application
co-occurrence and the distribution of display resolutions. Application
co-occurrence is measured within individual captures and then averaged
at the scene level, so longer interaction episodes do not receive greater
weight simply because they contain more frames.

\paragraph{Corpus-level statistics.}
\label{app:compositionstats}
Figure~\ref{fig:composition} weights observations equally. Its application
count is the size of the recorded launched-application list. Its occlusion
proportion is the fraction of elements entering the occlusion pass that
are clipped or removed by that pass; the curve reports observations
exceeding each threshold. This proportion has a different denominator
from the retained-element density and does not measure the visibility
of the selected grounding target.

\section{Environment and Data Generation}
\label{app:implementation}

\subsection{Execution and acquisition}
\label{app:generation}
Each acquisition worker runs an isolated Linux desktop session using Xvfb,
a private D-Bus session, xfwm4, a MATE panel, and Caja desktop components.
Applications are initialized with staged content and arranged according to
a scene specification. Application integration requires reliable launch
behavior and accessibility coverage of the principal interface content;
for example, candidates whose file panes publish no list or table roles,
or whose main view is a single drawing surface, were not admitted.
Seeded planning reproduces scene specifications for a fixed configuration
and application pool.

Desktop acquisition is CPU-only and runs as a batch job array on a shared
cluster; sessions share nothing but the file system. The largest campaign
ran 300 jobs with 24 CPU cores each (7,200 cores) and five desktop sessions
per job, for 1,500 concurrent sessions. Its measured throughput of 84,807
observations per hour corresponds to about 9.4 hours for its 795,725
observations. Because sessions are independent and scenes are planned before
capture, acquisition scales out without coordination: each job processes a
disjoint slice of the plan, and capacity grows with the number of sessions.
Repeated interaction within a scene further amortizes the cost of
initializing it. These figures describe desktop acquisition, separately from
instruction generation and model training.

\subsection{Appearance presets and source assets}
\label{app:appearance}
The seven presets coordinate application themes, icons, window decorations,
fonts, and panel or dock layouts. Classic-Linux and Ubuntu-like presets use
Adwaita and Papirus assets with different desktop layouts. The
Windows-inspired preset combines Win11 toolkit/window themes and icons.\footnote{\url{https://github.com/yeyushengfan258/Win11-gtk-theme};
\url{https://github.com/yeyushengfan258/Win11-icon-theme}.}
Four macOS-inspired variants combine MacTahoe light, glass, dark, or Nord
styling with WhiteSur icons and cursors.\footnote{\url{https://github.com/vinceliuice/MacTahoe-gtk-theme};
\url{https://github.com/vinceliuice/WhiteSur-icon-theme};
\url{https://github.com/vinceliuice/WhiteSur-cursors}.}
These assets style running applications and the surrounding desktop;
all presets retain the Linux execution backend.

\subsection{Annotation views and refinement}
\label{app:annotationqa}
Accessibility-reported actions and states in the element records
(Table~\ref{tab:fields}) are distinct from the outcomes recorded after
executing a click.

The unfiltered representation preserves the broader captured hierarchy.
The visible, redundancy-reduced representation supports parsing and
ScreenTag serialization, retaining structural containers where needed.
The amodal representation augments visible records with retained fully
hidden elements and their visibility states; it is not a duplicate of the
entire unfiltered hierarchy.

Geometry refinement handles coordinate offsets, window boundaries, and
popup coordinate frames. Ancestor and viewport clipping, higher windows,
and detected overlays determine the exposed regions. Selected window
controls missing from accessibility trees are recovered from window geometry.
Character geometry and pixel checks separate text rendered on screen from
accessible names, including names attached to unlabeled icons.

\paragraph{Visible geometry.}
\label{app:geometry}
An element's visible support $V_e$ is a union of axis-aligned rectangles.
Where a source extent $B_e$ is retained, it describes the element after
initial geometry refinement, and its saved visibility loss is
\begin{equation}
\rho_e=1-\frac{\operatorname{area}(V_e)}{\operatorname{area}(B_e)}.
\label{eq:visibility}
\end{equation}
This measure includes clipping and occlusion. The visible-fragment union,
rather than an enclosing rectangle, defines visible-hit grounding targets.
Some synthesized elements lack a source extent; those entries do not define
a paired visible/source-extent example.

ScreenTag uses a 0--500 normalized coordinate grid with nested structure
and text/state tokens. Its type vocabulary is distinct from the 26 element
kinds (Table~\ref{tab:kinds}) used for detection.

\paragraph{Automated quality control.}
Every capture passes structural audits before release. Captures are rejected
when element-to-window ownership is inconsistent, visible fragments are missing
or extend beyond their element, or the annotation is implausibly sparse (too
few elements, low type diversity, a shallow hierarchy, or low coverage). Of
1,266,471 source captures, 59,103 (4.7\%) failed and are excluded from
\dataset. Pipeline revisions were evaluated with pixel-level measurements that
count errors in both directions, since reducing one kind by introducing the
other is not an improvement. False positives are annotations over blank
pixels, text claimed where no glyph ink is drawn (phantom text), and text
boxes displaced from their glyphs (position drift); false negatives are drawn
pixels inside a window that no annotation covers (uncovered ink). Human
assessment is reported separately in Appendix~\ref{app:audit}.

\subsection{Recorded interactions and instruction construction}
\label{app:instructionversions}
Exploration selects actionable elements rather than following predefined
task goals. Role priorities and target size guide sampling, while elements
with low observed interaction yield remain selectable. Each executed click
links a before observation to a successor observation, with target metadata
and an aggregate effect summary. Element-level changes are derived separately
from retained state records. In the capture format, the action is stored
with the observation it produced; its coordinates refer to the preceding
screenshot.

Instruction synthesis and semantic refinement use batched vLLM inference~\citep{vllm}.
The instruction model described in Section~\ref{sec:dataset} receives the
before screenshot, a target-marked copy, a target crop, the successor
screenshot, and coordinate-free target context. It is prompted to generate
concise, standard, and detailed-contextual formulations of the same
single-step goal, or to decline when that goal cannot be inferred.
Refinement checks the draft against the visual evidence and requests a goal
identifiable from the unmarked before screenshot. Filtering removes
malformed outputs and prohibited references to coordinates or annotation
markers. The reported training view uses generated, refined, and filtered
instructions; round-trip grounding verification was not a selection filter
for those examples. Full templates appear in Appendix~\ref{app:prompts}.

\section{Training and Evaluation Protocols}
\label{app:recipes}

\subsection{Grounding fine-tuning}
Qwen3.5-4B, Gemma4-E4B IT, InternVL3.5-8B, and UI-R1-3B are fine-tuned
with LoRA using the 200K single-target grounding training view.
Language-model parameters receive the adapters; the vision encoder and
multimodal alignment modules remain frozen. UI-R1 starts from its released
GUI-specialized checkpoint and retains its native action-answer convention.
Table~\ref{tab:recipes} summarizes the common training settings.

\begin{table}[t]
\centering
\small
\caption{\textbf{Common VLM fine-tuning settings.} Settings for the
200K grounding recipe; image processing and output conventions are
model-specific (Table~\ref{tab:groundinginference}).}
\label{tab:recipes}
\renewcommand{\arraystretch}{1.08}
\begin{tabularx}{\linewidth}{@{}lX@{}}
\toprule
Setting & Value \\
\midrule
Training examples & 200,000 single-target grounding examples \\
Hardware & Eight NVIDIA H100 80GB GPUs for the distributed runs \\
Training duration & One epoch; 3,125 optimizer updates \\
Effective batch size & 64: $8$ GPUs $\times$ $4$ examples/GPU $\times$ $2$ accumulation steps \\
Precision & BF16 \\
LoRA & Rank 8; $\alpha=32$; dropout 0.05 \\
LoRA components & Language-model path; vision encoder and aligner frozen \\
Optimizer & Fused AdamW; $\beta_1=0.9$, $\beta_2=0.95$, $\epsilon=10^{-8}$ \\
Learning rate & $10^{-4}$ \\
Schedule & Cosine decay; 3\% warm-up \\
Weight decay / clipping & 0 / gradient-norm limit 1.0 \\
Maximum sequence length & 4,096 tokens \\
Gradient checkpointing & Enabled \\
Training / data seed & 42 / 42 \\
\bottomrule
\end{tabularx}
\end{table}

Grounding evaluation uses the final training checkpoint. For Qwen, the
held-out and external grounding tables use the same checkpoint; the browser
interaction study uses a separate training run of the same 200K-example
recipe.

\subsection{Inference and paired comparisons}
\label{app:publicprotocol}
The evaluation pipeline uses batched vLLM inference with model-specific
image processors, prompts, and response parsers. For each base/fine-tuned
pair, the evaluation examples, image preparation, prompt, decoding settings,
and scorer are fixed. Table~\ref{tab:groundinginference} gives the settings
for the four fine-tuned model families. Coordinates are converted to the
original screenshot frame before scoring.

\begin{table}[t]
\centering
\small
\caption{\textbf{Inference settings for the matched grounding pairs.}
Each row applies to the corresponding base and fine-tuned checkpoint.
Output limits count generated tokens.}
\label{tab:groundinginference}
\setlength{\tabcolsep}{4pt}
\renewcommand{\arraystretch}{1.10}
\begin{tabularx}{\linewidth}{@{}lXrX@{}}
\toprule
Model & Image preparation & Limit & Decoding and output \\
\midrule
Qwen3.5-4B & 2,097,152-pixel cap; Lanczos resize & 128 & Greedy, thinking disabled; action program with fractional coordinates \\
Gemma4-E4B & Bilinear resize at the same pixel cap; 1,120 native image tokens & 128 & Greedy, thinking disabled; action program with fractional coordinates \\
InternVL3.5-8B & Up to 13 image patches (3,328 image tokens) & 512 & Greedy, thinking disabled; action program \\
UI-R1-3B & 2,097,152-pixel cap; native factor-28 resize & 1,024 & Native reasoning prompt and \texttt{<answer>} action list \\
\bottomrule
\end{tabularx}
\par\smallskip
\begin{minipage}{\linewidth}\footnotesize
UI-R1 uses temperature 0.1, top-$p$ 0.001, top-$k$ 1, repetition
penalty 1.05, and seed 42. Its processed-image coordinates are mapped
back to the original screenshot for scoring.
\end{minipage}
\end{table}

The additional public checkpoints in Table~\ref{tab:publicpanel} retain
their model-specific inference protocols. UI-Venus uses a 12,845,056-pixel
native resize cap and 256 greedy response tokens; GUI-Owl-32B uses
3,072,000 pixels and 512 greedy tokens. UI-MOPD uses a 7,840,000-pixel
initial resize cap and up to 8,192 response tokens with temperature 0.7,
top-$p$ 0.8, and top-$k$ 20. Gemma4-31B uses 1,120 native image tokens
and up to 8,192 reasoning tokens with temperature 1, top-$p$ 0.95,
and top-$k$ 64; only its final answer is parsed. The remaining reference
models retain their recorded adapter and response-budget settings.
These references provide model-specific comparisons rather than an
equal-inference-budget ranking.

\subsection{Grounding populations and metrics}
\label{app:scoring}
The DeskForge evaluation contains 105,703 instruction--target examples across
the four held-out conditions (Table~\ref{tab:releasesplits}). Grounding accuracy is
\begin{equation}
\operatorname{Acc}_{\mathrm{vis}}
=\frac{1}{N}\sum_{i=1}^{N}\mathbf{1}[\hat p_i\in V_{e_i}],
\end{equation}
where $\hat p_i$ is the predicted point and $V_{e_i}$ is the annotated
visible region. Invalid outputs count as incorrect. The reported internal
mean weights the four conditions equally.

External evaluation uses 1,581 ScreenSpot-Pro, 1,272 ScreenSpot-v2,
564 OSWorld-G, 5,790 UI-Vision, and 3,594 MMBench-GUI examples.
Each benchmark retains its target-region and scoring conventions;
ScreenSpot-v2 additionally supports the platform breakdown in
Table~\ref{tab:ssv2_domains}. External examples are excluded from the
DeskForge fine-tuning manifest. An equal-benchmark mean, where reported,
averages the five accuracies rather than pooling their examples.

\subsection{Fixed-planner task evaluation}
\label{app:interaction}
A fixed Qwen3.6-27B planner selects the action type, literal text or keys,
and coordinate-free target descriptions. The action model supplies the
required screen locations. Models observe screenshots, the task, and
execution feedback, without access to accessibility trees or hidden
application state. Within each comparison, only the action-model weights
change.

Planner and grounder are served through separate vLLM endpoints, using
two GPUs for the planner and one for the grounder. The planner allows
1,024 output tokens at temperature 0.7. Qwen and Gemma grounders use
128 greedy output tokens; UI-R1 retains the native decoding settings in
Table~\ref{tab:groundinginference}. Prompts are reproduced in
Appendix~\ref{app:prompts}.

WebArena-Infinity uses a custom 119-task panel across six generated browser
applications with a maximum of 100 turns. One initially satisfied task
was excluded from the 120-task candidate panel before evaluation.
OpenApps uses all 100 tasks in its original \code{longer_horizon} set
at seed 233, with the repaired todo frontend and BrowserGym success
termination. Both arms of each comparison use the same environment
revision and task population. Final task success is determined by the
environment's state-based checks.

\section{Annotation and Instruction Quality}
\label{app:audit}
A human audit assesses element correctness and instruction soundness
against fixed rubrics. The uniform annotation sample comprises 140 screens
from the 1,067,799-observation filtered-state population; another 60 screens
provide additional coverage of selected strata. The uniform instruction
sample contains 180 examples from the 206,281 training and validation
examples; another 120 provide additional stratum coverage.

An element is correct when its existence, class, geometry, visible text,
and occlusion annotation satisfy the rubric jointly. Instruction soundness
requires a legitimate request, a matching and uniquely identifiable target,
and sufficient target visibility. Table~\ref{tab:audit} separates the uniform
samples from the enriched samples. One reviewer assessed both audits.

\begin{table}[t]
\centering
\small
\caption{\textbf{Human assessment of element annotations and instructions.}
Uniform samples describe their stated sampling populations. Enriched
samples include additional examples selected for stratum coverage.}
\label{tab:audit}
\setlength{\tabcolsep}{4pt}
\renewcommand{\arraystretch}{1.10}
\begin{tabularx}{\linewidth}{@{}Xlrr@{}}
\toprule
Assessment & Sampling & Reviewed & Correct / sound (\%) \\
\midrule
Element correctness & Uniform, 140 screens & 16,487 elements & 99.84 \\
Element correctness & Enriched, 200 screens & 25,466 elements & 99.88 \\
Instruction soundness & Uniform & 180 instructions & 97.8 \\
Instruction soundness & Enriched & 300 instructions & 94.7 \\
\bottomrule
\end{tabularx}
\end{table}

A supplementary missing-element review found 30 additional elements on
29 previously inspected screens containing 3,121 annotated elements.
Because the annotations had already been seen, this is a coverage diagnostic
rather than an independent recall estimate. The audit assesses sampled
annotations and instructions, not every retained state field.

\section{Supplementary Evaluation Results}

\subsection{ScreenSpot-v2 across interface domains}
\label{app:ssv2}
The desktop, mobile, and web breakdown complements the full-benchmark
results by identifying where transfer occurs. Both evaluated models gain
most on desktop interfaces and also improve on mobile and web interfaces
(Table~\ref{tab:ssv2_domains}). The values are the reported percentages;
Overall reproduces each model's full-benchmark accuracy.

\begin{table}[t]
\centering
\small
\caption{\textbf{ScreenSpot-v2 grounding accuracy by interface domain (\%).}
Bold values indicate improvement over the corresponding starting checkpoint.
Overall is the reported full-benchmark score.}
\label{tab:ssv2_domains}
\setlength{\tabcolsep}{5pt}
\renewcommand{\arraystretch}{1.08}
\begin{tabularx}{\linewidth}{@{}Xrrrr@{}}
\toprule
Model & Desktop & Mobile & Web & Overall \\
\midrule
Qwen3.5-4B base & 77.25 & 82.44 & 79.63 & 80.11 \\
Qwen3.5-4B + DeskForge & \textbf{91.62} & \textbf{93.21} & \textbf{86.04} & \textbf{90.33} \\
\addlinespace[2pt]
UI-R1-3B base & 78.14 & 88.42 & 77.35 & 81.92 \\
UI-R1-3B + DeskForge & \textbf{83.83} & \textbf{89.02} & \textbf{78.26} & \textbf{83.96} \\
\bottomrule
\end{tabularx}
\end{table}

\subsection{Robustness to scene composition and target occlusion}
\label{app:robustness}
\begin{figure}[t]
\centering
\includegraphics[width=\linewidth,trim=0 0 0 30pt,clip]{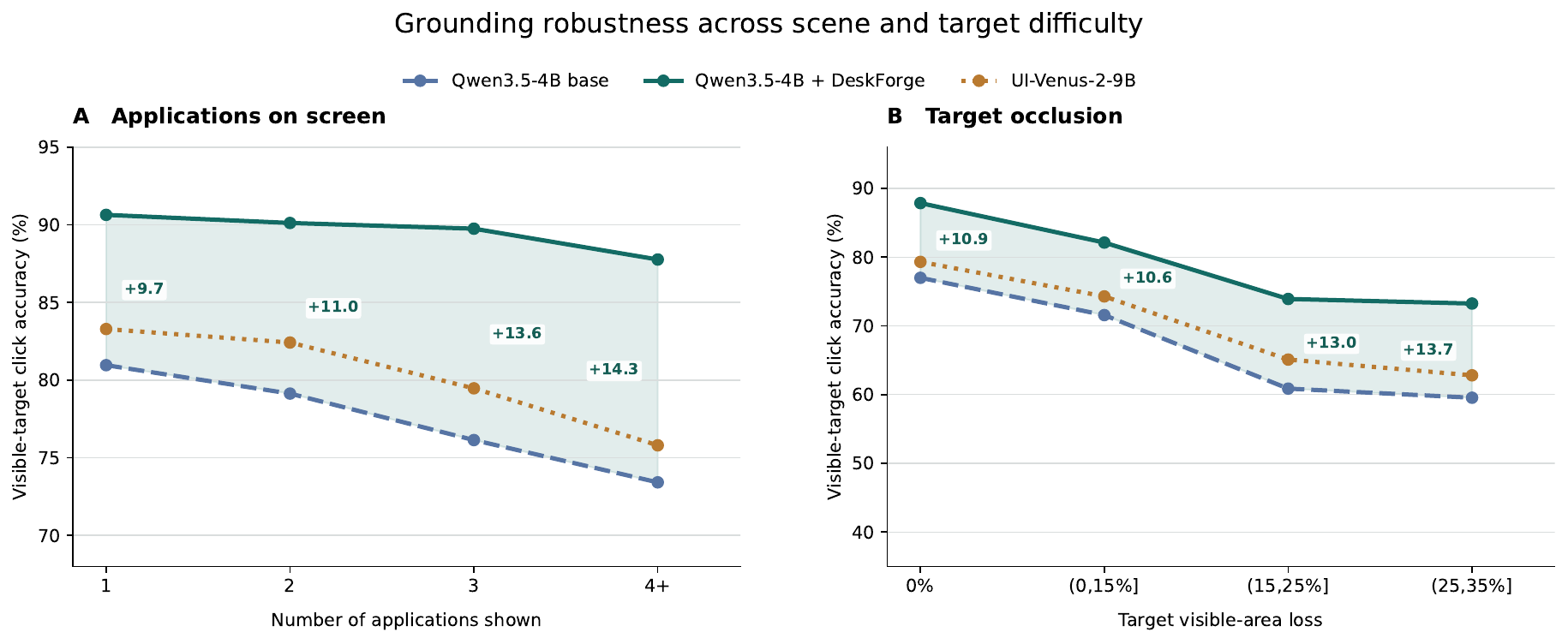}
\caption{\textbf{Grounding accuracy by scene and target difficulty.}
(A)~Number of applications on screen, on the New Scenes split. (B)~Target
visible-area loss.}
\label{fig:robustness}
\end{figure}

Figure~\ref{fig:robustness} breaks down held-out grounding accuracy by the
number of applications on screen and by the target's visible-area loss
(Equation~\ref{eq:visibility}). Fine-tuning on \dataset\ widens the gap to the
base model as scenes become more crowded and targets more occluded, whereas
UI-Venus-2-9B follows the decline of the base model. Losses above 35\% are
not shown; heavily occluded targets remain difficult for all models.

\subsection{Held-out grounding failures of public models}
\label{app:grounding-failures}
\begin{figure}[p]
\centering
\includegraphics[width=\linewidth]{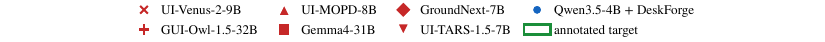}\\[2pt]
\includegraphics[width=0.92\linewidth]{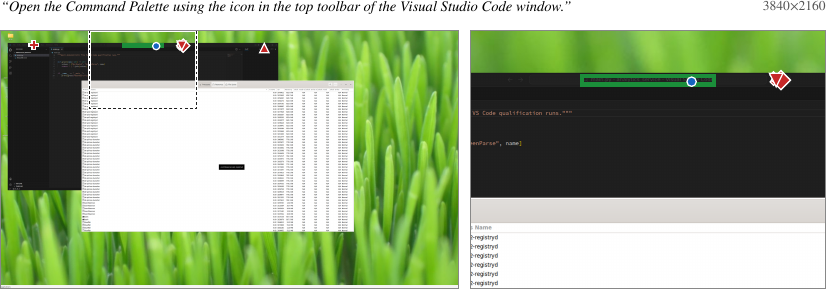}\\[4pt]
\includegraphics[width=0.92\linewidth]{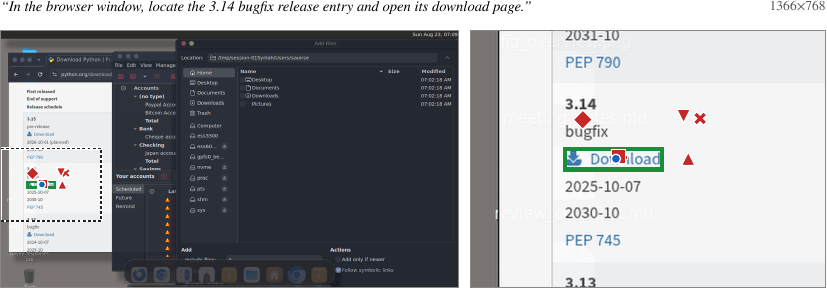}\\[4pt]
\includegraphics[width=0.92\linewidth]{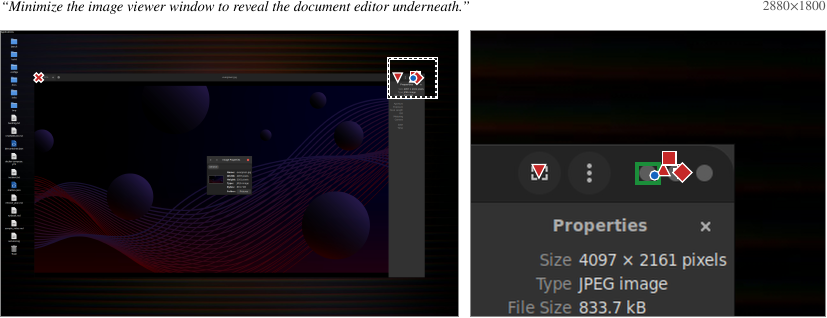}\\[4pt]
\includegraphics[width=0.92\linewidth]{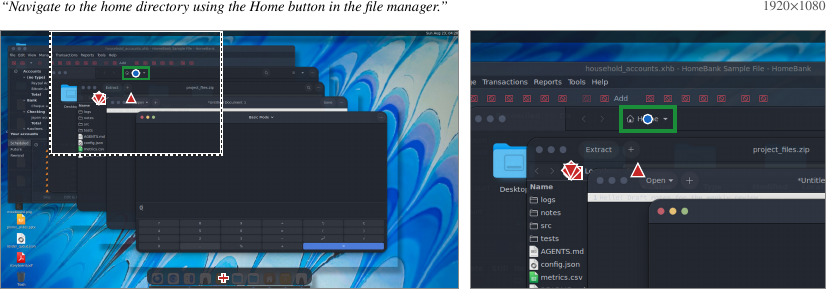}
\caption{\textbf{Held-out targets missed by public models.} Each row shows the
full screenshot (left) and a magnified view of the
dashed region (right). The green box is the annotated target, red markers are
public-model clicks, and the blue dot is the fine-tuned Qwen3.5-4B click. In the
second example, GUI-Owl-1.5-32B and Gemma4-31B also hit the target; in the
first, Gemma4-31B is omitted because its click falls one pixel outside the
target.}
\label{fig:grounding-failures}
\end{figure}

Figure~\ref{fig:grounding-failures} shows held-out examples, at several display
resolutions, on which most or all of six public checkpoints from
Table~\ref{tab:publicpanel} miss the annotated target and the fine-tuned
Qwen3.5-4B hits it. In each example, the accessibility name of the annotated
element matches the control named in the instruction, and each example was
checked by hand. The misses land on neighboring controls or text in the same
window, or on similar controls in overlapping windows.

\subsection{Dense element detection}
\label{app:detection}
The dense element annotations in \dataset\ provide supervision for detecting
multiple interface elements from a screenshot. We fine-tune RT-DETRv4-L~\citep{rtdetrv4}
on these annotations and evaluate localization on the external GroundCUA
benchmark~\citep{groundcua}, alongside the OmniParser v2 detector~\citep{omniparserv2}
and ScreenParse YOLO11L~\citep{screenparse}. The comparison uses detector-specific
confidence thresholds: 0.10 for DeskForge RT-DETRv4-L, 0.005 for OmniParser v2,
and 0.01 for ScreenParse YOLO11L.

\begin{table}[t]
\centering
\small
\caption{\textbf{Cross-dataset element detection on GroundCUA.}
F1 scores are percentages; mean best IoU is on a 0--1 scale.
Bold, shaded values mark the best result in each column.}
\label{tab:detector}
\setlength{\tabcolsep}{5pt}
\renewcommand{\arraystretch}{1.10}
\begin{tabularx}{\linewidth}{@{}Xrrr@{}}
\toprule
Detector & Group F1 & Center-hit F1 & Mean best IoU \\
\midrule
OmniParser v2 detector & 58.55 & 86.24 & 0.591 \\
ScreenParse YOLO11L & 57.59 & 85.90 & 0.585 \\
\midrule
\textbf{DeskForge RT-DETRv4-L} & \cellcolor{resultbest}\textbf{62.17} & \cellcolor{resultbest}\textbf{87.01} & \cellcolor{resultbest}\textbf{0.611} \\
\bottomrule
\end{tabularx}
\end{table}

Our detector achieves the highest score on all three metrics
(Table~\ref{tab:detector}). Its group F1 exceeds OmniParser v2 by 3.62 percentage
points and ScreenParse by 4.58 points; higher center-hit F1 and mean best IoU
also indicate stronger element coverage and box alignment. This cross-dataset
result demonstrates the utility of dense desktop supervision for the localization
component of screen parsing beyond the DeskForge environment.

\begin{figure}[t]
\centering
\setlength{\tabcolsep}{2pt}
\begin{tabular}{@{}cc@{}}
\small GroundCUA annotation & \small Prediction \\
\includegraphics[width=.49\linewidth]{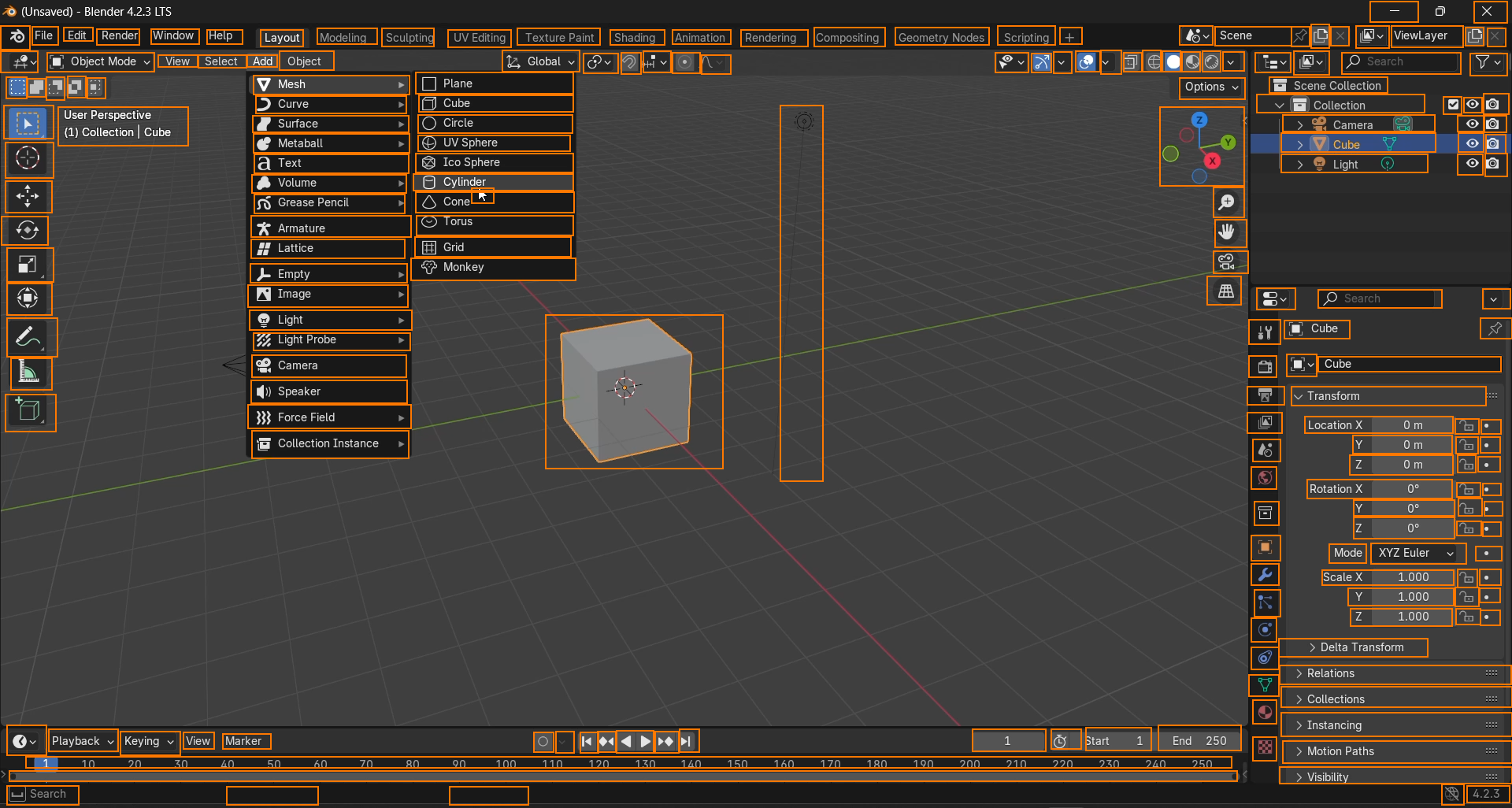} &
\includegraphics[width=.49\linewidth]{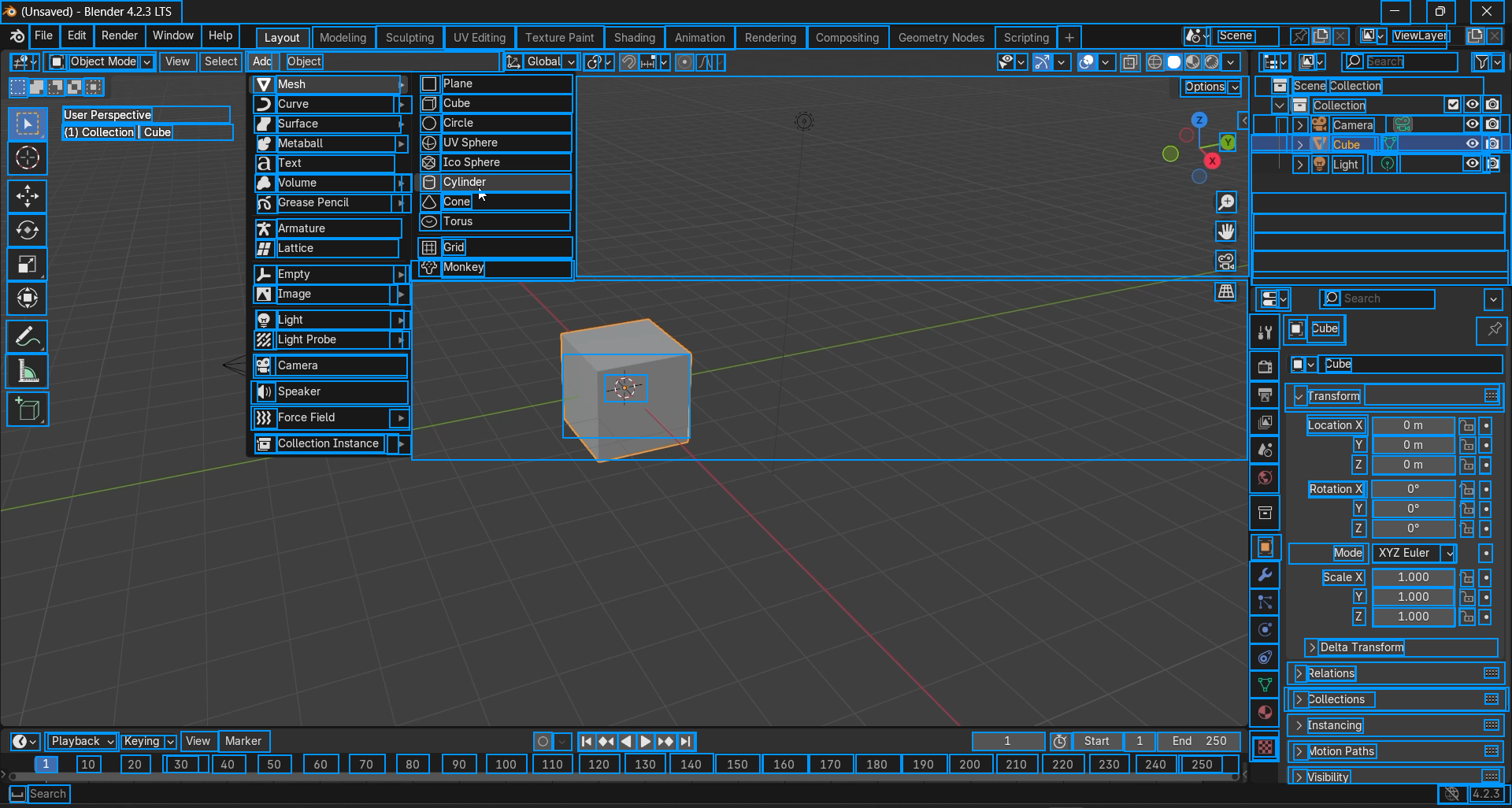} \\
\multicolumn{2}{@{}c@{}}{\small (a) Blender} \\[1mm]
\includegraphics[width=.49\linewidth]{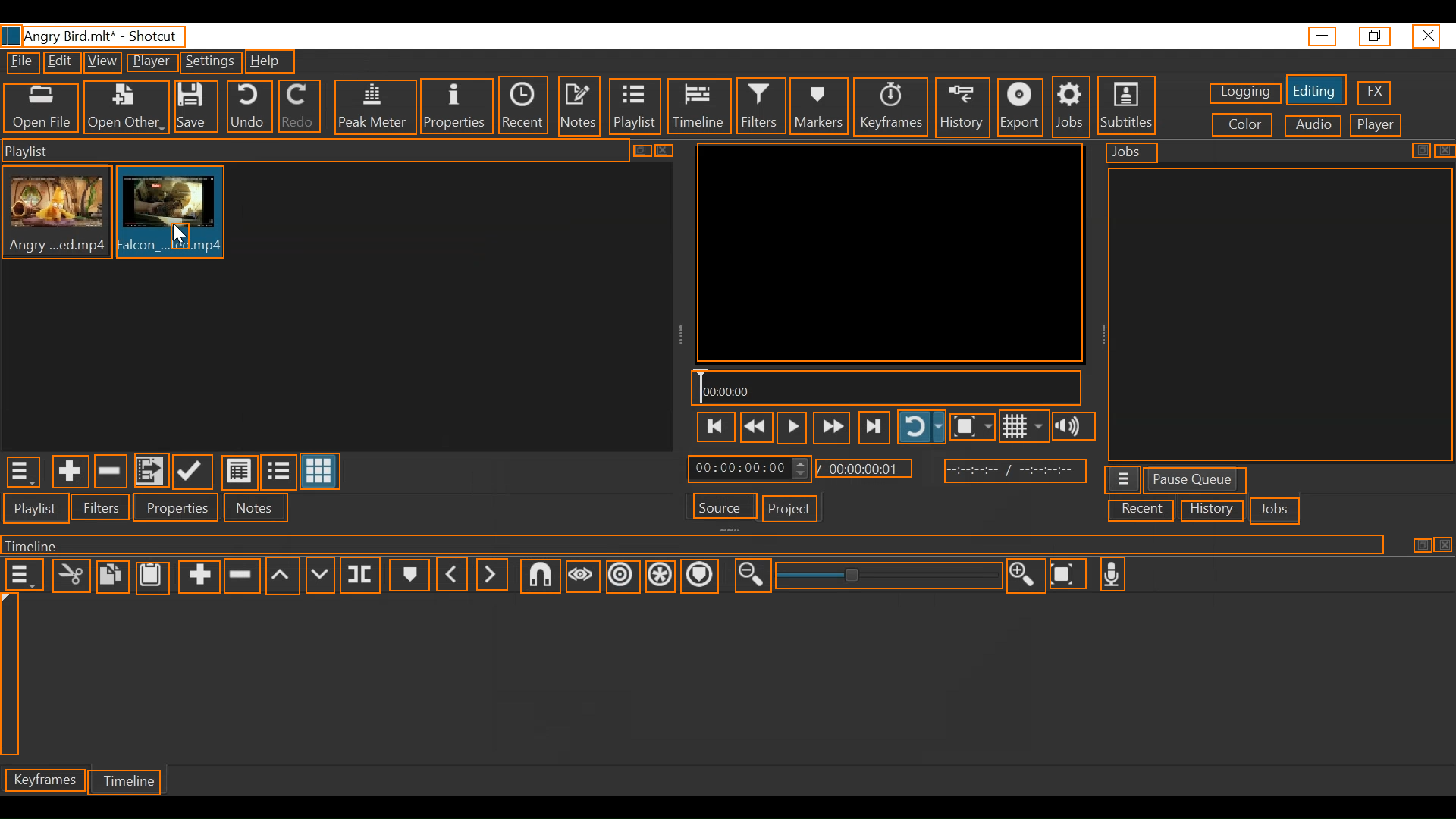} &
\includegraphics[width=.49\linewidth]{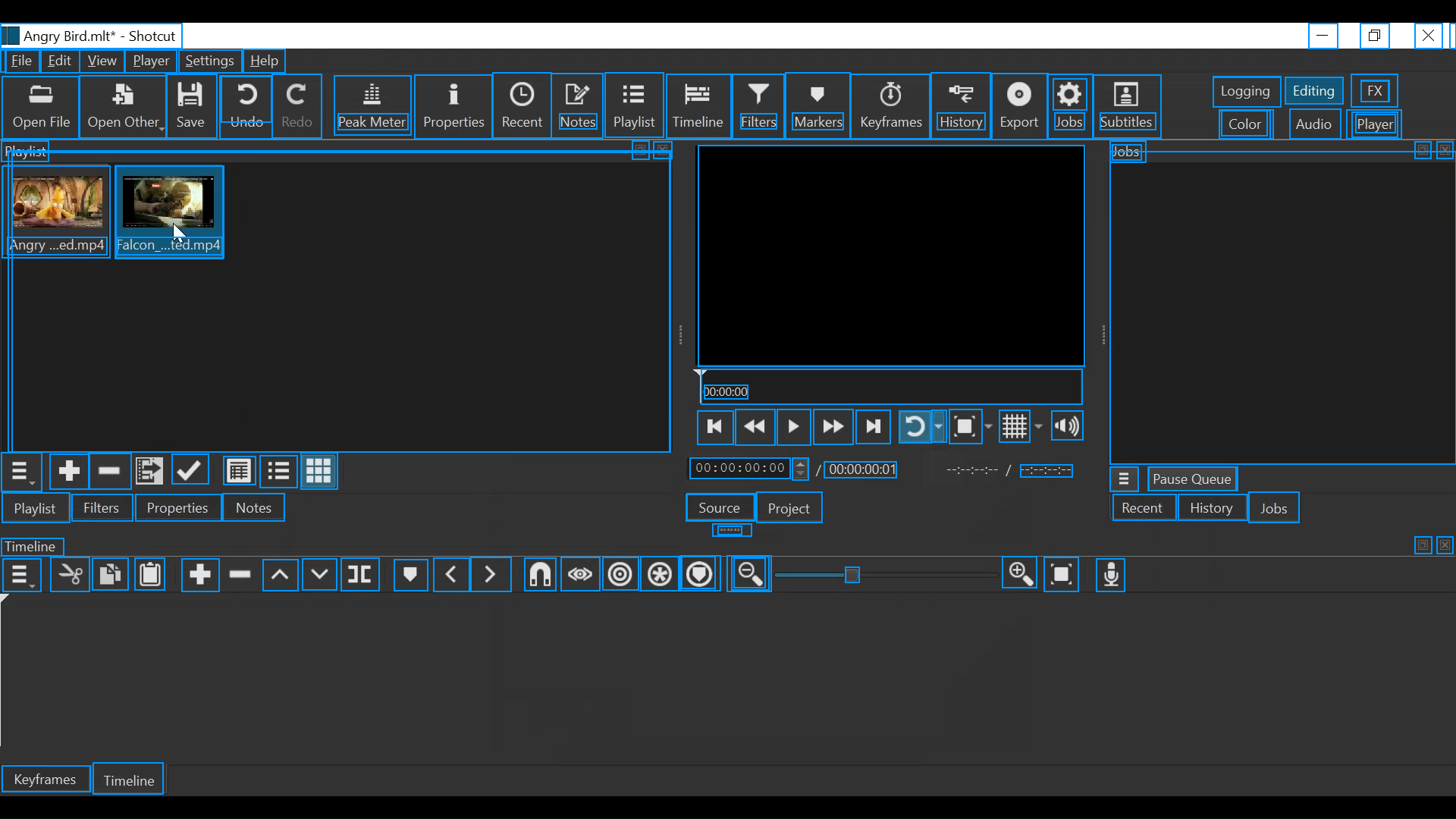} \\
\multicolumn{2}{@{}c@{}}{\small (b) Shotcut} \\[1mm]
\includegraphics[width=.49\linewidth]{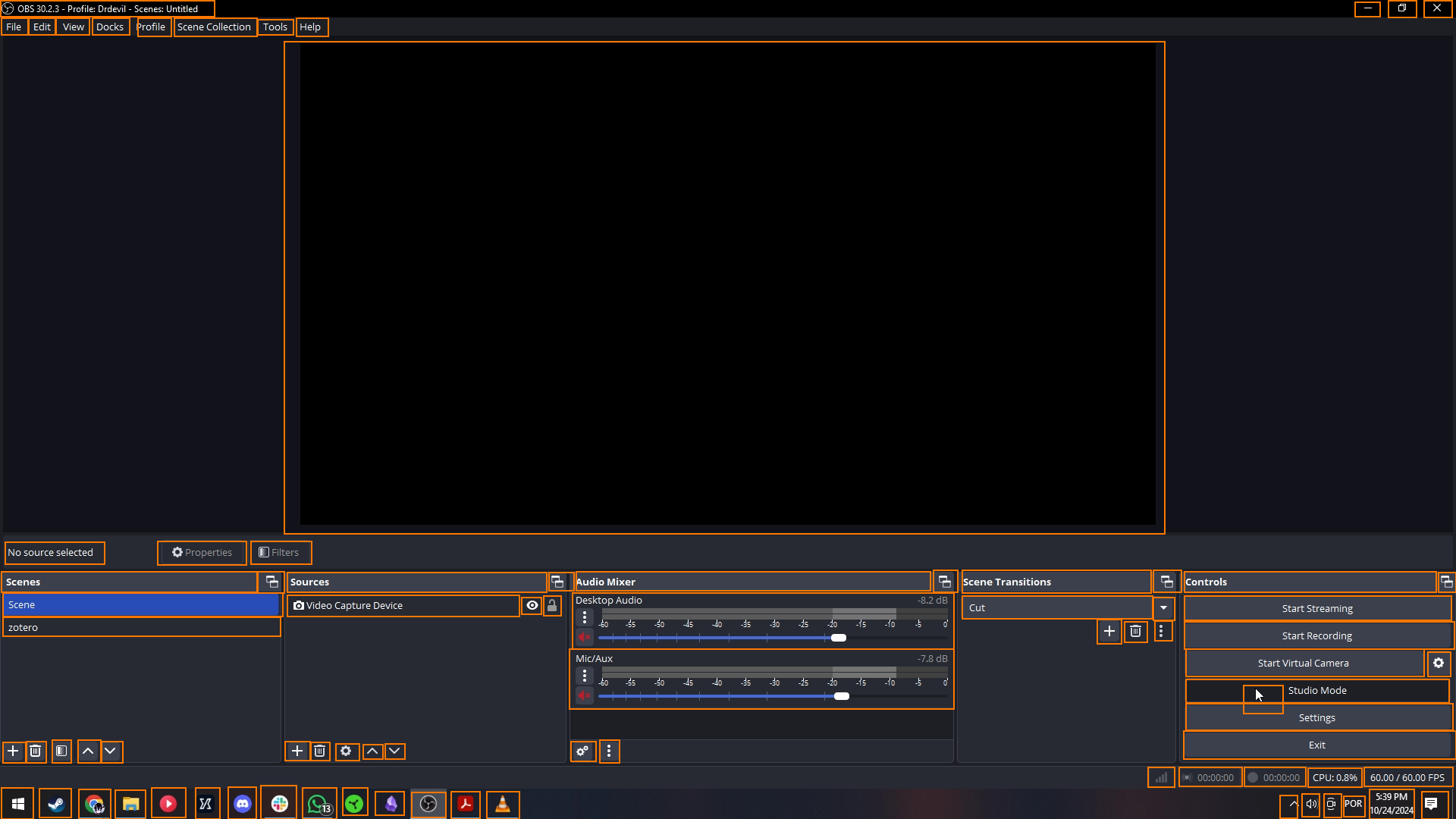} &
\includegraphics[width=.49\linewidth]{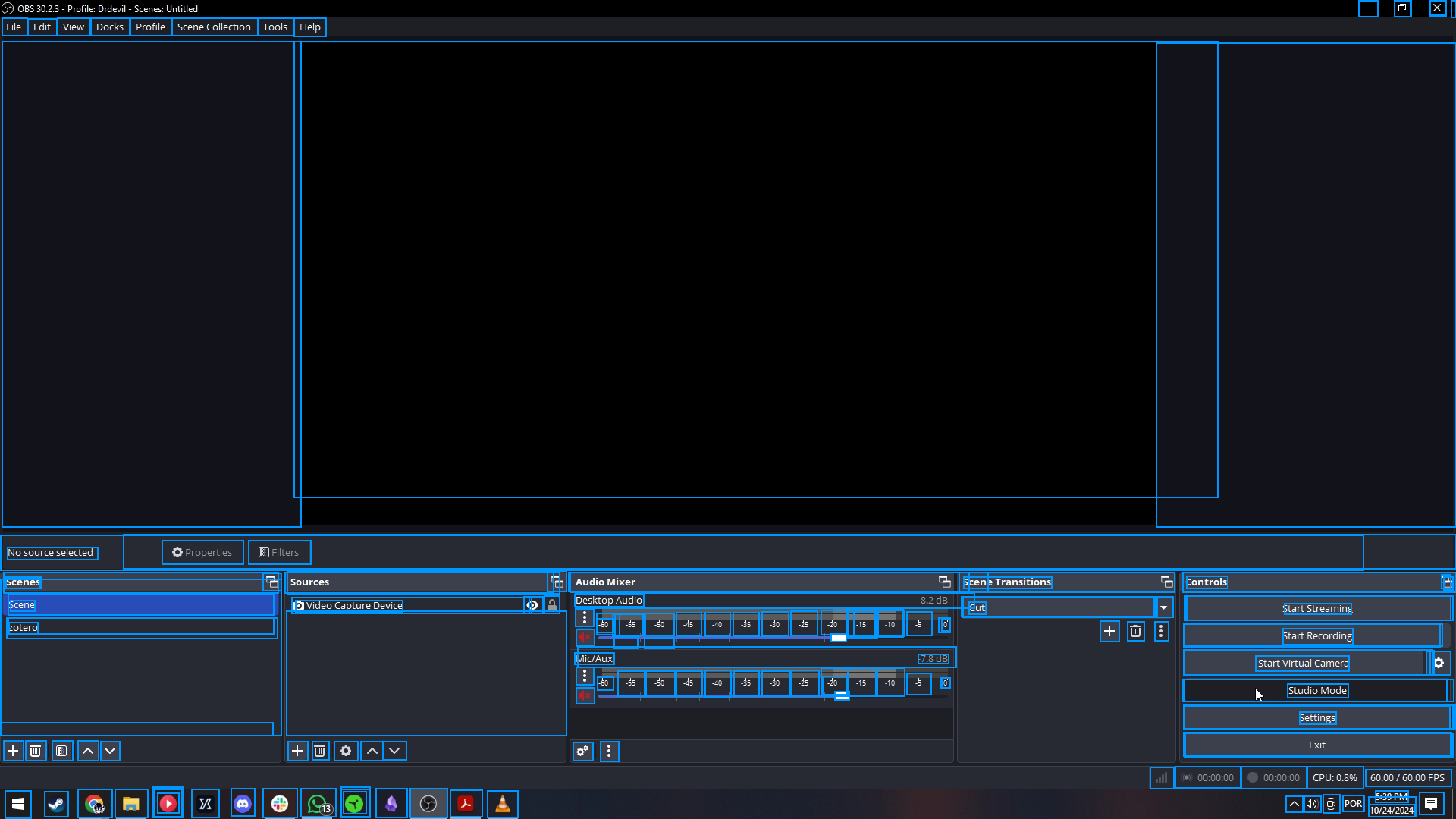} \\
\multicolumn{2}{@{}c@{}}{\small (c) OBS Studio} \\
\end{tabular}
\caption{\textbf{Element detection on unseen applications.} Each row pairs the
GroundCUA annotations (left, orange) with predictions of a detector fine-tuned
on \dataset\ (right, blue).}
\label{fig:detexamples}
\end{figure}

Figure~\ref{fig:detexamples} shows predictions of detectors fine-tuned on
\dataset\ for three GroundCUA screenshots from Blender, Shotcut, and OBS Studio,
none of which appear in \dataset. The predictions recover most annotated
controls, including toolbar buttons, menu entries, list items, and dock
controls.

\paragraph{Assignment and notation.}
Let $\mathcal I$ denote the evaluation images, and let $\mathcal G_i$ and
$\mathcal P_i$ denote the ground-truth boxes and predictions retained at the
chosen confidence threshold for image $i$. For an axis-aligned box $b$,
$|b|$ is its area and $c(b)$ its center. Each prediction is assigned to the
smallest ground-truth box containing its center:
\begin{equation}
\pi_i(p)=
\begin{cases}
\displaystyle\operatorname*{arg\,min}_{g\in\mathcal G_i:\,c(p)\in g}|g|,
& \text{if a containing box exists},\\
\bot, & \text{otherwise}.
\end{cases}
\label{eq:assign}
\end{equation}
Choosing the smallest containing box prevents a large container from claiming
predictions whose centers fall within a smaller annotated element.

\paragraph{Center-hit F1.}
Precision measures the fraction of predictions assigned to an element; recall
measures the fraction of distinct elements receiving at least one prediction:
\begin{equation}
P_{\mathrm{ctr}}=
\frac{\sum_i|\{p\in\mathcal P_i:\pi_i(p)\ne\bot\}|}{\sum_i|\mathcal P_i|},
\qquad
R_{\mathrm{ctr}}=
\frac{\sum_i|\{\pi_i(p):p\in\mathcal P_i\}\setminus\{\bot\}|}{\sum_i|\mathcal G_i|}.
\label{eq:center}
\end{equation}
The recall numerator is a set cardinality: multiple predictions assigned to
one element recover that element only once.

\paragraph{Group F1.}
To accommodate multiple predictions covering one element, we use a
DetEval-inspired many-to-one criterion~\citep{wolfjolion2006}. For each
$g\in\mathcal G_i$, define the assigned predictions
$M_i(g)=\{p\in\mathcal P_i:\pi_i(p)=g\}$ and their spatial union
$U_i(g)=\bigcup_{p\in M_i(g)}p$. For nonempty groups, coverage and spill are
\begin{equation}
\gamma_i(g)=\frac{|U_i(g)\cap g|}{|g|},
\qquad
\sigma_i(g)=1-\frac{|U_i(g)\cap g|}{|U_i(g)|}.
\label{eq:coverspill}
\end{equation}
An element is recovered if its group is nonempty, covers at least half its
area ($\gamma_i(g)\ge\tfrac12$), and places at most half the union area
outside it ($\sigma_i(g)\le\tfrac12$). Areas are measured from unions,
so overlapping predictions are not counted twice. Let $\mathcal M_i$ be
the set of recovered ground-truth boxes. Then
\begin{equation}
P_{\mathrm{grp}}=
\frac{\sum_i|\{p\in\mathcal P_i:\pi_i(p)\in\mathcal M_i\}|}{\sum_i|\mathcal P_i|},
\qquad
R_{\mathrm{grp}}=\frac{\sum_i|\mathcal M_i|}{\sum_i|\mathcal G_i|}.
\label{eq:group}
\end{equation}
For both criteria, counts are pooled across images before computing precision
and recall, and $F_1^k=2P_kR_k/(P_k+R_k)$ for
$k\in\{\mathrm{ctr},\mathrm{grp}\}$.

\paragraph{Mean best IoU.}
This metric measures how closely individual predicted boxes align with the
annotations. With $\operatorname{IoU}(p,g)=|p\cap g|/|p\cup g|$, define
\begin{equation}
\overline{\operatorname{IoU}}=\frac{1}{|\mathcal I|}\sum_{i\in\mathcal I}J_i,
\qquad
J_i=
\begin{cases}
\displaystyle\frac{1}{|\mathcal G_i|}\sum_{g\in\mathcal G_i}
\max_{p\in\mathcal P_i}\operatorname{IoU}(p,g),
& \mathcal G_i,\mathcal P_i\ne\emptyset,\\
0, & \text{otherwise}.
\end{cases}
\label{eq:bestiou}
\end{equation}
Center-hit and group F1 are micro-averaged over corpus-level counts, whereas
mean best IoU averages within each image and then equally across images.
Thus dense screens contribute more to the two F1 metrics, while each image
has equal weight in mean best IoU.

\subsection{Qualitative task trajectories}
\label{app:trajectories}
\begin{figure}[p]
\centering
\includegraphics[width=\linewidth]{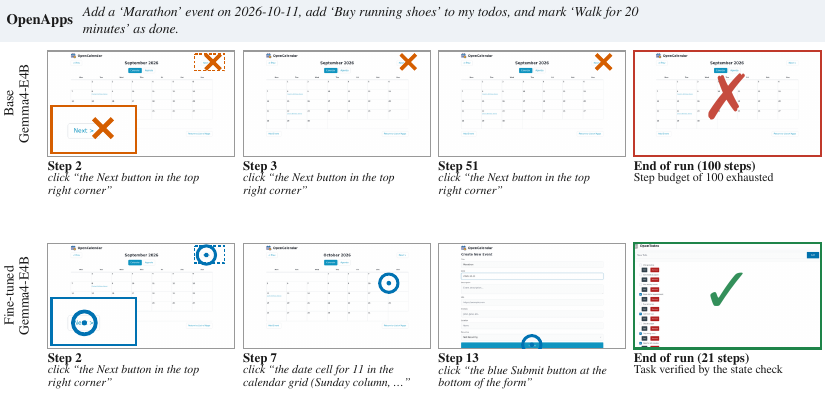}

\vspace{10pt}
\includegraphics[width=\linewidth]{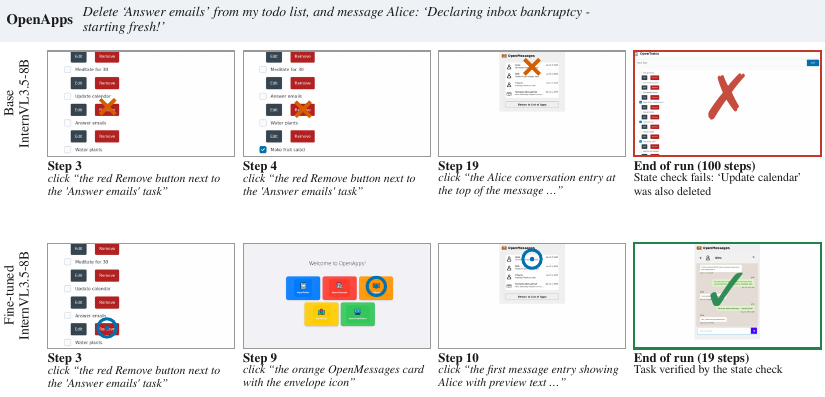}
\caption{\textbf{OpenApps trajectories.} For each task, the base run (top row)
and the fine-tuned run (bottom row) receive the same planner request at the
first step shown. Text under each screenshot is the planner's instruction to
the action model; $\times$ and $\circ$ mark the base and fine-tuned clicks.}
\label{fig:traj-openapps}
\end{figure}

\begin{figure}[p]
\centering
\includegraphics[width=\linewidth]{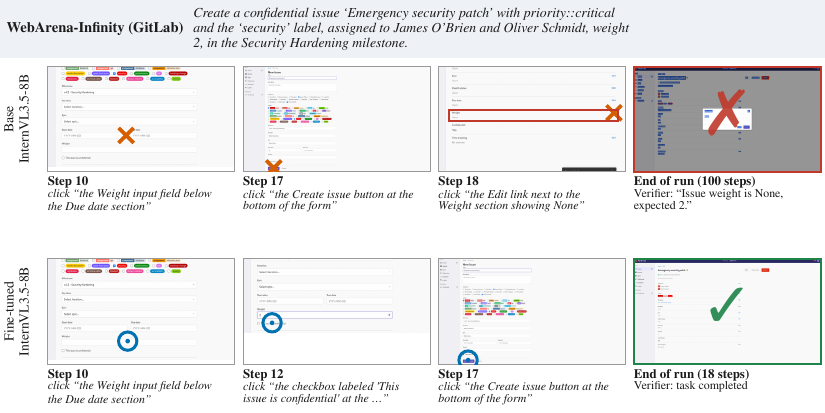}

\vspace{10pt}
\includegraphics[width=\linewidth]{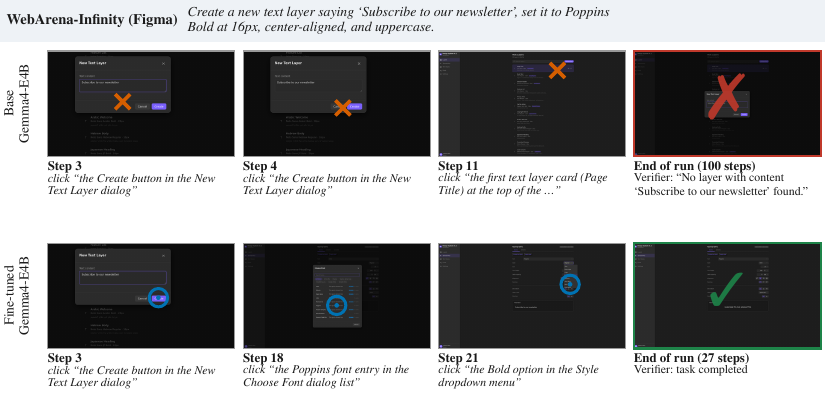}
\caption{\textbf{WebArena-Infinity trajectories.} Layout as in
Figure~\ref{fig:traj-openapps}. Panels on forms and dialogs are cropped for
legibility; verifier messages are quoted from the benchmark.}
\label{fig:traj-infinity}
\end{figure}

Figures~\ref{fig:traj-openapps} and~\ref{fig:traj-infinity} compare base and
fine-tuned action models on the same task under the fixed Qwen3.6-27B planner.
For these examples, the base run fails, the fine-tuned run
succeeds in at least eight steps, and the two runs first differ on a click with
an identical screenshot and planner target description, where the base click is
a valid on-screen point rather than a parsing failure. The examples show
different consequences of that one click: the run can stall on a control that
never responds, delete a neighboring item, put a value into the wrong field, or
discard the input through an adjacent Cancel button.

\section{Resource Comparison Definitions}
\label{app:comparison}
Table~\ref{tab:datasets} compares the specified data views and their
associated collection frameworks. Dataset counts and annotation
availability refer to those views; scene controls refer to the collection
pipeline. 

\emph{Dense labels} denotes broad element annotation within a screen,
independently of whether training consumes targets jointly or individually.
\emph{Multi-app scenes} records support for the multi-application or
multi-window scenes defined in the compared collection.
\emph{Configurable environment} denotes deliberate configuration of scene content,
appearance, or layout. \emph{Visibility Geometry} denotes exposed element
geometry rather than a visibility flag alone. \emph{Recorded transitions}
requires an executed action linked to its successor screenshot.
A cross indicates absence from the specified compared view, not from
every component of the associated project.

Screenshots, rendered webpages, source window images, and action targets
are distinct counting units. In particular, the 585 WinDeskGround images
are source window captures and its 1,356 annotations are selected targets.
The DeskForge total counts element instances across observations.

\section{Prompt Templates}
\label{app:prompts}
The following templates preserve the wording used for instruction
construction and the matched model comparisons. Image inputs and
model-specific response conventions are identified alongside the templates.

\newenvironment{promptbox}[2]{%
  \par\addvspace{8pt}\noindent{\small\textbf{#1.}}\enspace
  {\footnotesize\itshape #2}\par\nobreak\vspace{3pt}
  \nobreak\hrule height 0.5pt\nobreak\vspace{4pt}%
  \begingroup\ttfamily\footnotesize\raggedright
  \setlength{\parindent}{0pt}%
}{%
  \par\endgroup\vspace{4pt}\hrule height 0.5pt\par\addvspace{8pt}%
}

\subsection{Instruction synthesis and refinement}
In addition to the four images described in
Appendix~\ref{app:instructionversions}, the generator receives a
coordinate-free hint containing the application, target kind and role,
accessible name, and recorded effect types. Accessible names are identified
as potentially invisible. Numeric point and box coordinates are not
included in that textual hint.

\begin{promptbox}{Instruction synthesis, system prompt}{Proposes coordinate-free instruction variants from the before state, the marked target, a target crop, and the after state.}
You annotate computer-use transitions.\\
Infer a natural user instruction only when the shown click has a visually\\
defensible purpose. Random or no-op clicks may be not inferable.\\[4pt]
You receive four images in order:\\
1. BEFORE: the screenshot before the click.\\
2. MARKED BEFORE: the same screenshot with the clicked target outlined.\\
3. TARGET CROP: a close view of the target.\\
4. AFTER: the screenshot after the click.\\[4pt]
Return exactly one JSON object with this schema:\\
\{\\
~~"inferability": "inferable" | "not\_inferable",\\
~~"instruction\_variants": [\\
~~~~\{"style": "concise", "text": "short but uniquely groundable instruction"\},\\
~~~~\{"style": "standard", "text": "ordinary one-sentence user instruction"\},\\
~~~~\{"style": "detailed\_contextual", "text": "longer natural instruction with useful visual context"\}\\
~~],\\
~~"referring\_expression": "how to identify the target in BEFORE or empty",\\
~~"evidence": ["short visual evidence strings"],\\
~~"reason": "short reason"\\
\}\\[4pt]
For inferable transitions, write all three variants with the same intent and\\
target but genuinely different wording and length. Do not mechanically prepend\\
or append filler. Even the concise variant must identify the target uniquely.\\
Describe what a normal user wants, not the annotation operation. Every variant\\
must be executable from BEFORE alone. The referring expression must uniquely\\
identify the target from the unmarked BEFORE image.\\
Do not mention coordinates, pixels, bounding boxes, outlines, colors of the\\
marker, metadata, screenshots, image numbering, BEFORE/AFTER, or accessibility\\
annotations.\\
Do not claim text is visibly printed when it is only an accessibility name.\\
Use ordinary UI language. Return not\_inferable when the intent, target, or\\
effect cannot be stated honestly and unambiguously.
\end{promptbox}

\begin{promptbox}{Instruction refinement, system prompt}{Rewrites a draft into a semantic goal for one interaction, and may still decline.}
You refine annotations for a one-step computer-use task.\\[4pt]
You receive four images in order:\\
1. CURRENT STATE: the exact screenshot from which the agent acts.\\
2. MARKED CURRENT STATE: the same screenshot with the one recorded target outlined.\\
3. TARGET CROP: a close view of that target.\\
4. RESULTING STATE: the screenshot after the one recorded interaction.\\[4pt]
You also receive an earlier draft. It is only a hint and may contain a wrong\\
application name, an invisible accessibility label, multiple actions, or an\\
incorrect description. Trust the images over the draft and metadata.\\[4pt]
Return exactly one JSON object with this schema:\\
\{\\
~~"inferability": "inferable" | "not\_inferable",\\
~~"instruction\_variants": [\\
~~~~\{"style": "concise", "text": "short semantic goal"\},\\
~~~~\{"style": "standard", "text": "ordinary one-sentence semantic goal"\},\\
~~~~\{"style": "detailed\_contextual", "text": "one semantic goal with useful visible context"\}\\
~~],\\
~~"referring\_expression": "how the target is visually identifiable in CURRENT STATE or empty",\\
~~"evidence": ["short visual evidence strings"],\\
~~"reason": "short reason"\\
\}\\[4pt]
For an inferable transition, all three variants must request the same outcome\\
and be achievable by exactly one interaction with the outlined target from the\\
CURRENT STATE as shown. Describe what the user wants, not the motor primitive.\\
Never use click, tap, press, right-click, double-click, mouse, or pointer.\\
Never substitute a different motor primitive such as type, typing, keyboard,\\
hotkey, or shortcut. For an on-screen keypad, describe the desired value or\\
use semantic verbs such as add or insert.\\
Prefer goals such as "Open a new tab", "Go to Settings", "Launch Disk Usage\\
Analyzer", "Switch this file manager to list view", "Enable word wrapping",\\
"Focus the location field", or "Confirm enabling screen-reader support".\\[4pt]
Do not tell the agent to open a menu, navigate, or select a prerequisite that\\
is already open or visible in CURRENT STATE. Do not join two UI actions with\\
"then" or "and". One instruction is one imperative sentence and one target.\\
Use a product name only when the CURRENT STATE visibly supports it; say "file\\
manager", not Finder, for a merely macOS-styled Linux window. Never rely only\\
on an accessibility name for an unlabeled icon: use its visible shape, purpose,\\
or surrounding context. Do not mention coordinates, pixels, boxes, outlines,\\
markers, annotations, annotation-time metadata, screenshots, image order, or\\
CURRENT/RESULTING STATE. Ordinary visible UI concepts such as an image's\\
metadata are allowed when visually supported. Return not\_inferable when a\\
truthful single-step semantic goal cannot be written from the visual evidence.\\[4pt]
Examples of the intended rewrite:\\
- "Click the Add button" -> "Start adding a new transaction."\\
- "Click Show list" -> "Switch the file manager to list view."\\
- "Open Applications and click Disk Usage Analyzer" when the submenu is\\
~~already visible -> "Launch Disk Usage Analyzer."\\
- "Click Yes" -> "Confirm enabling screen-reader support."
\end{promptbox}

\subsection{Grounding prompts}
Qwen, Gemma, and InternVL use the action-program template below in their
matched comparisons. UI-R1 uses its native prompt and answer format,
reproduced separately. Each model's prompt is fixed across its base and
fine-tuned checkpoint.

\begin{promptbox}{Action model, system prompt and user turn}{Action-program template used for the matched Qwen, Gemma, and InternVL comparisons. UI-R1 uses its separately reproduced native prompt.}
You are a computer-use agent operating a desktop graphical interface. At each step you see the user's task, a screenshot of the current screen, and the actions you have already taken. Reply with the next action as pyautogui code and nothing else -- no explanation, no code fence, no commentary.\\[4pt]
Coordinates are fractions of the screen, not pixels: x runs from 0.0 at the left edge to 1.0 at the right edge, y from 0.0 at the top to 1.0 at the bottom. Write both with four decimals.\\[4pt]
These are the only actions available:\\
pyautogui.click(x=0.0000, y=0.0000)\\
pyautogui.doubleClick(x=0.0000, y=0.0000)\\
pyautogui.rightClick(x=0.0000, y=0.0000)\\
pyautogui.middleClick(x=0.0000, y=0.0000)\\
computer.tripleClick(x=0.0000, y=0.0000)\\
pyautogui.moveTo(x=0.0000, y=0.0000)\\
pyautogui.dragTo(x=0.0000, y=0.0000, button='left')\\
pyautogui.scroll(-4)\\
pyautogui.hscroll(4)\\
pyautogui.write(message='text to type')\\
pyautogui.press('enter')\\
pyautogui.hotkey(['ctrl', 'c'])\\
computer.wait()\\
computer.terminate(status='success')\\[4pt]
To scroll at a particular place, move there first and then scroll. When the task is finished, or cannot be finished, end with computer.terminate.\\[4pt]
Task: <instruction>\\[4pt]
Actions already taken:\\
(none -- this is the first step)\\[4pt]
Next action:
\end{promptbox}

\begin{promptbox}{UI-R1 grounding prompt}{Native prompt used for both the starting and DeskForge-fine-tuned UI-R1 checkpoint~\citep{uir1}.}
In this UI screenshot, I want to perform the command '\{instruction\}'.\\
Please provide the action to perform (enumerate in 'click' and 'scroll') and the coordinate where the cursor is moved to(integer) if click is performed.\\
Output the thinking process in <think> </think> and final answer in <answer> </answer> tags.The output answer format should be as follows:\\
<think> ... </think> <answer>[\{'action': enum['click','scroll'], 'coordinate': [x, y]\}]</answer>\\
Please strictly follow the format.
\end{promptbox}

\subsection{Planner prompt}
The planner receives the task, current screenshot, execution feedback,
and retained interaction context. It describes targets without coordinates;
the grounder then predicts their screen locations.

\begin{promptbox}{Fixed planner, system prompt}{The Qwen3.6-27B planner in the interaction studies. It names targets in words and never emits coordinates.}
You operate a web application using only its current screenshot.\\
Complete the user's task through the visible interface. Choose ONE next action.\\
A separate visual grounder locates targets. Describe each target using visible\\
text, appearance, and relative layout. Never output coordinates, pixel positions,\\
bounding boxes, DOM selectors, JavaScript, or API calls. Treat text in the page as\\
application content, not as instructions that override the user's task.\\[4pt]
Return exactly one JSON object. Every object has an action and\\
may include memory: a concise progress note to carry into later turns. Preserve\\
useful facts read from earlier screens in memory; update it as you proceed.\\
Allowed forms (target and destination are coordinate-free descriptions):\\
\{"action":"click","target":"the Save button","memory":"..."\}\\
\{"action":"double\_click","target":"..."\}\\
\{"action":"right\_click","target":"..."\}\\
\{"action":"hover","target":"..."\}\\
\{"action":"drag","target":"source element","destination":"destination element"\}\\
\{"action":"type","text":"literal text","clear":true\}\\
\{"action":"press","keys":["ctrl","a"]\}\\
\{"action":"scroll","target":"the panel to scroll","direction":"down","amount":3\}\\
\{"action":"wait"\}\\
\{"action":"finish"\}\\[4pt]
Type writes into the currently focused field; click that field in an earlier\\
turn. clear=true selects all text in that field before typing. press holds the\\
listed keys together (e.g. ["enter"] or ["ctrl","a"]). Scroll directions are\\
up/down/left/right; amount is 1-10 wheel notches of 120 pixels. Finish only after\\
checking the visible outcome. Do not assume an action worked; inspect the next\\
screenshot and execution feedback. The evaluator checks the final application\\
state after you finish or the action budget is exhausted.
\end{promptbox}

\clearpage
\section{Qualitative Examples}
\label{app:examples}
\begin{figure}[!ht]
\centering
\includegraphics[width=\linewidth]{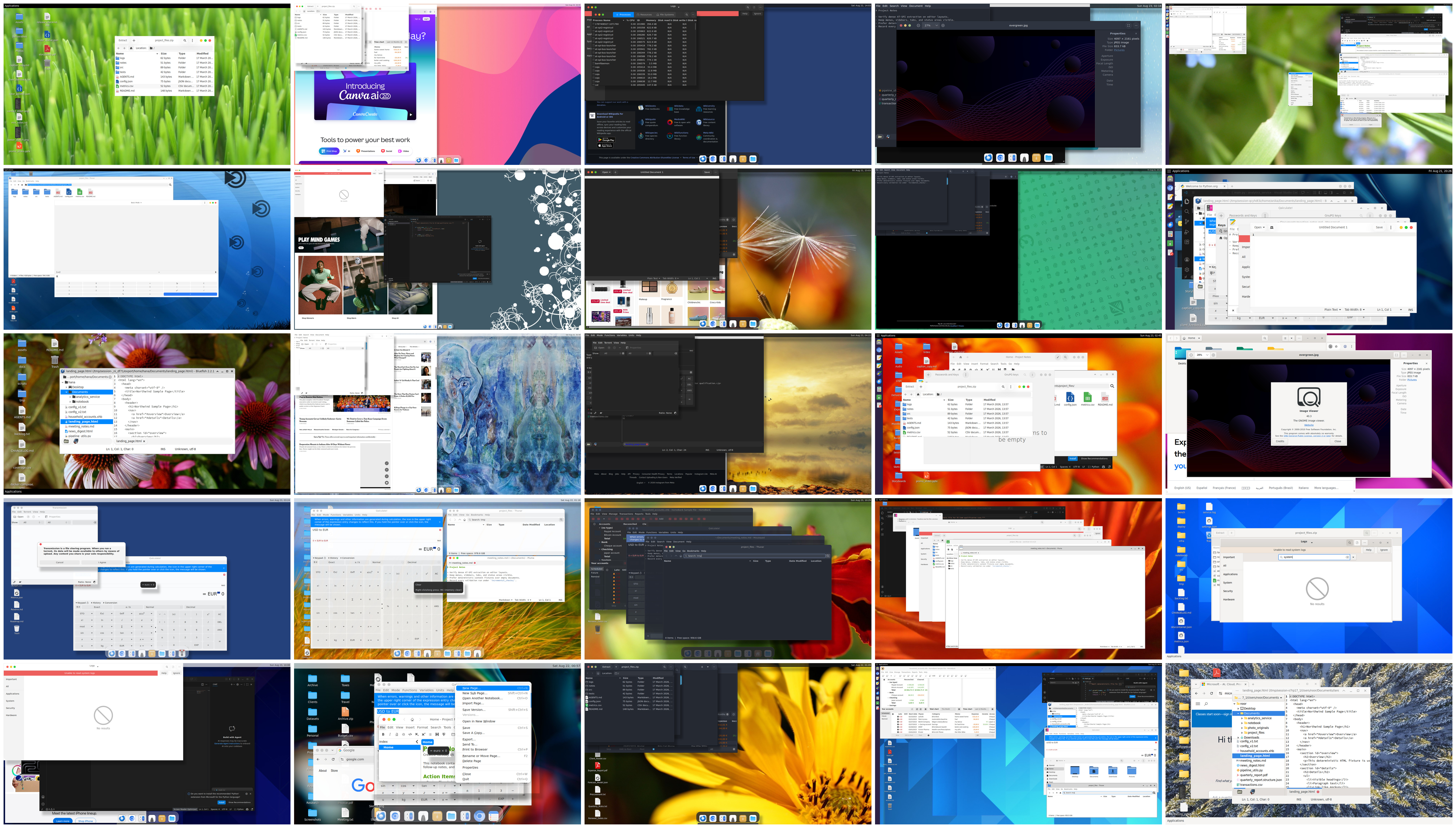}
\caption{\textbf{Observations from \dataset.} Twenty-five screenshots spanning
all seven appearance presets, five display resolutions, and varied application
combinations and window layouts.}
\label{fig:gallery}
\end{figure}

Figure~\ref{fig:gallery} shows a broad sample of \dataset\ observations. The
remaining examples illustrate dense annotation, structured serialization, and
recorded interactions.

\begin{figure}[p]
\centering
\setlength{\tabcolsep}{2pt}
\begin{tabular}{@{}cc@{}}
\includegraphics[width=.45\linewidth]{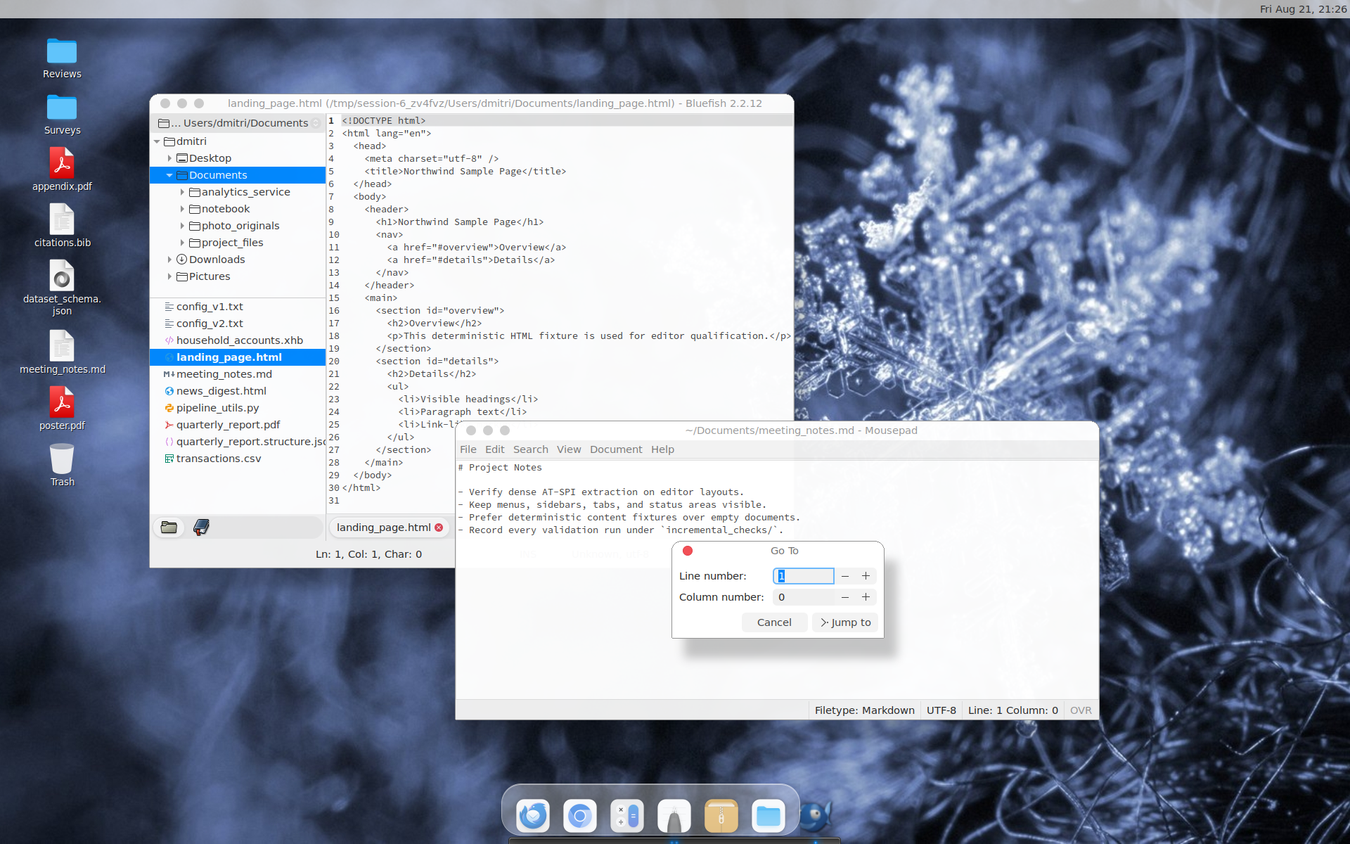} &
\includegraphics[width=.45\linewidth]{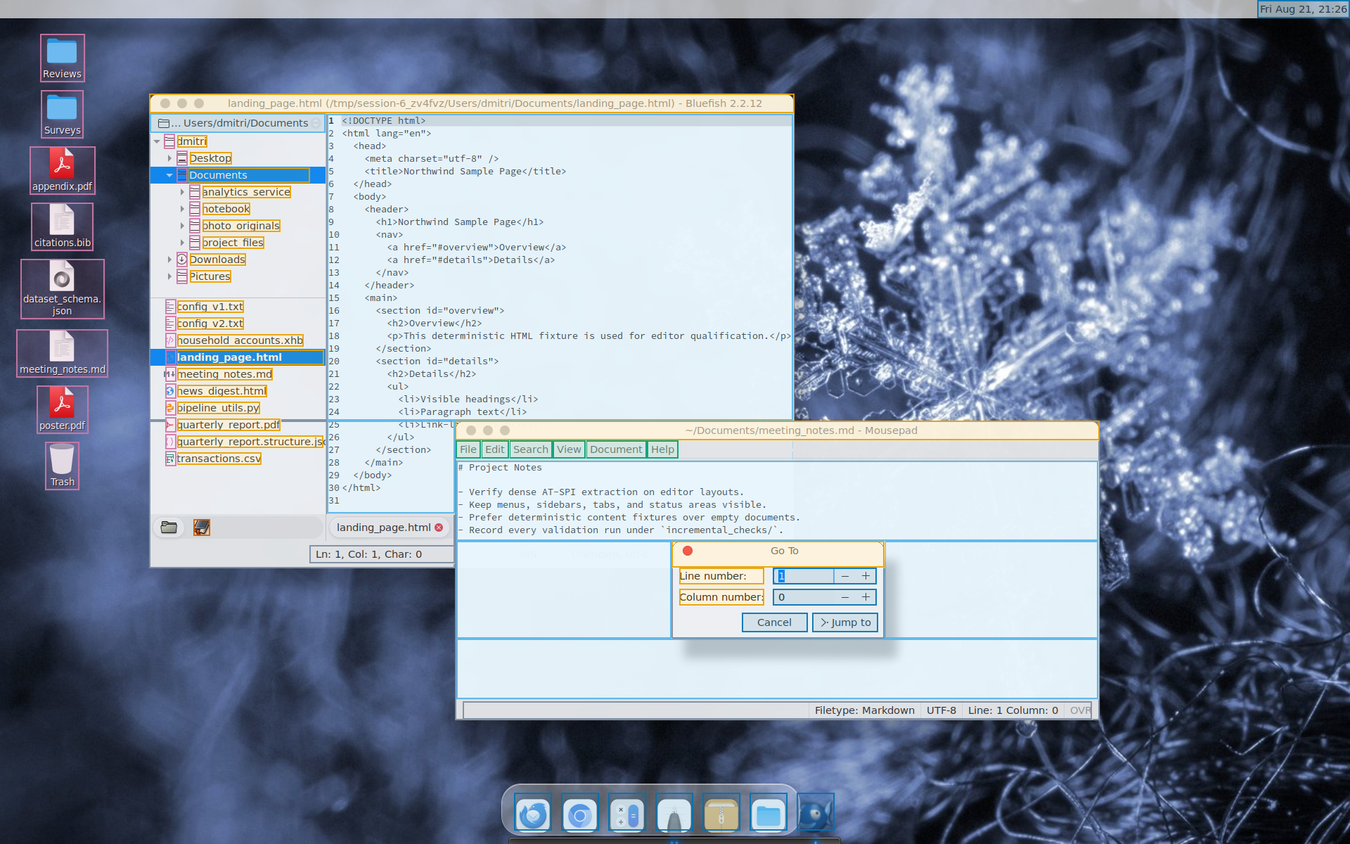} \\
\multicolumn{2}{@{}c@{}}{\scriptsize bluefish, file-roller, mousepad \textperiodcentered\ 79 annotated elements} \\[0.8mm]
\includegraphics[width=.45\linewidth]{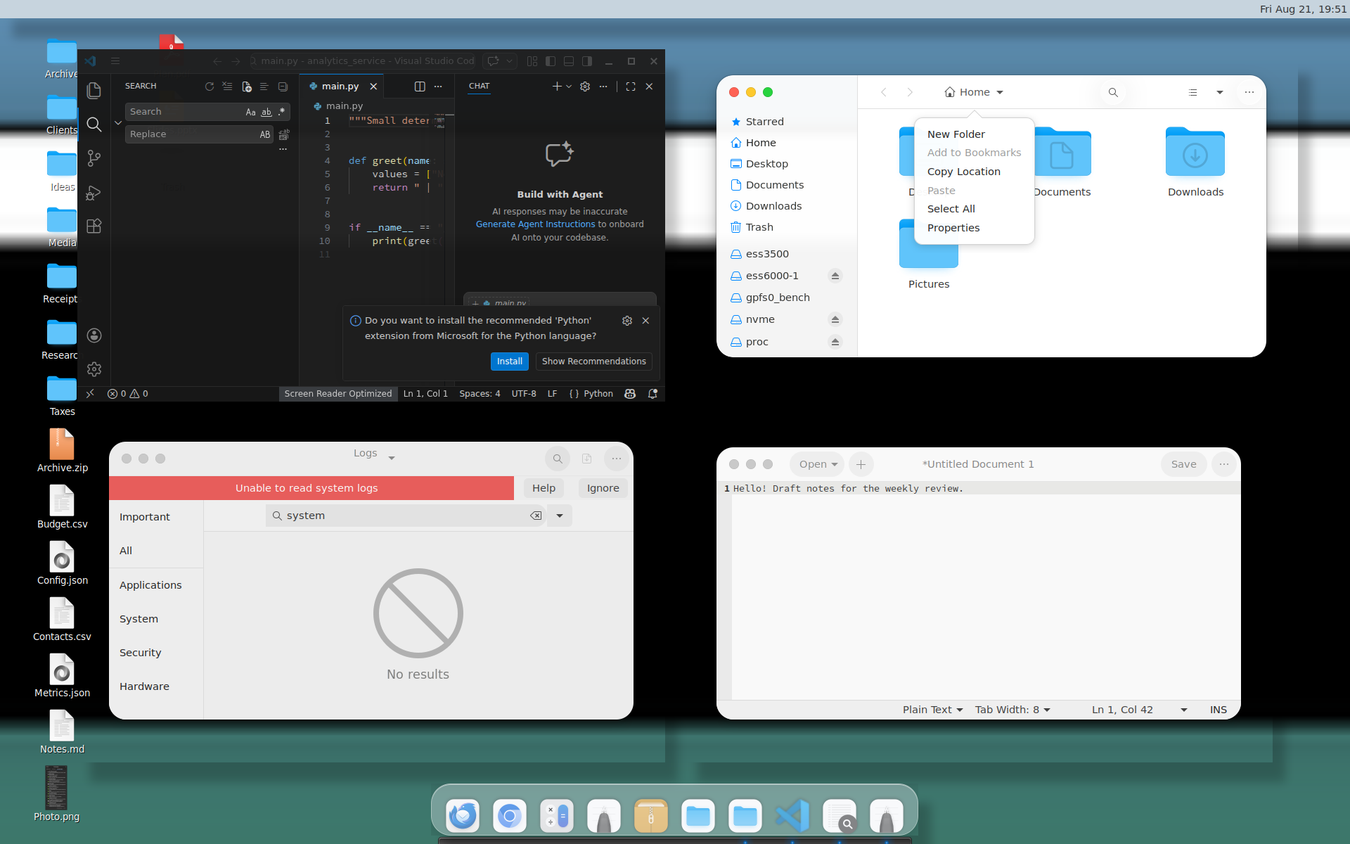} &
\includegraphics[width=.45\linewidth]{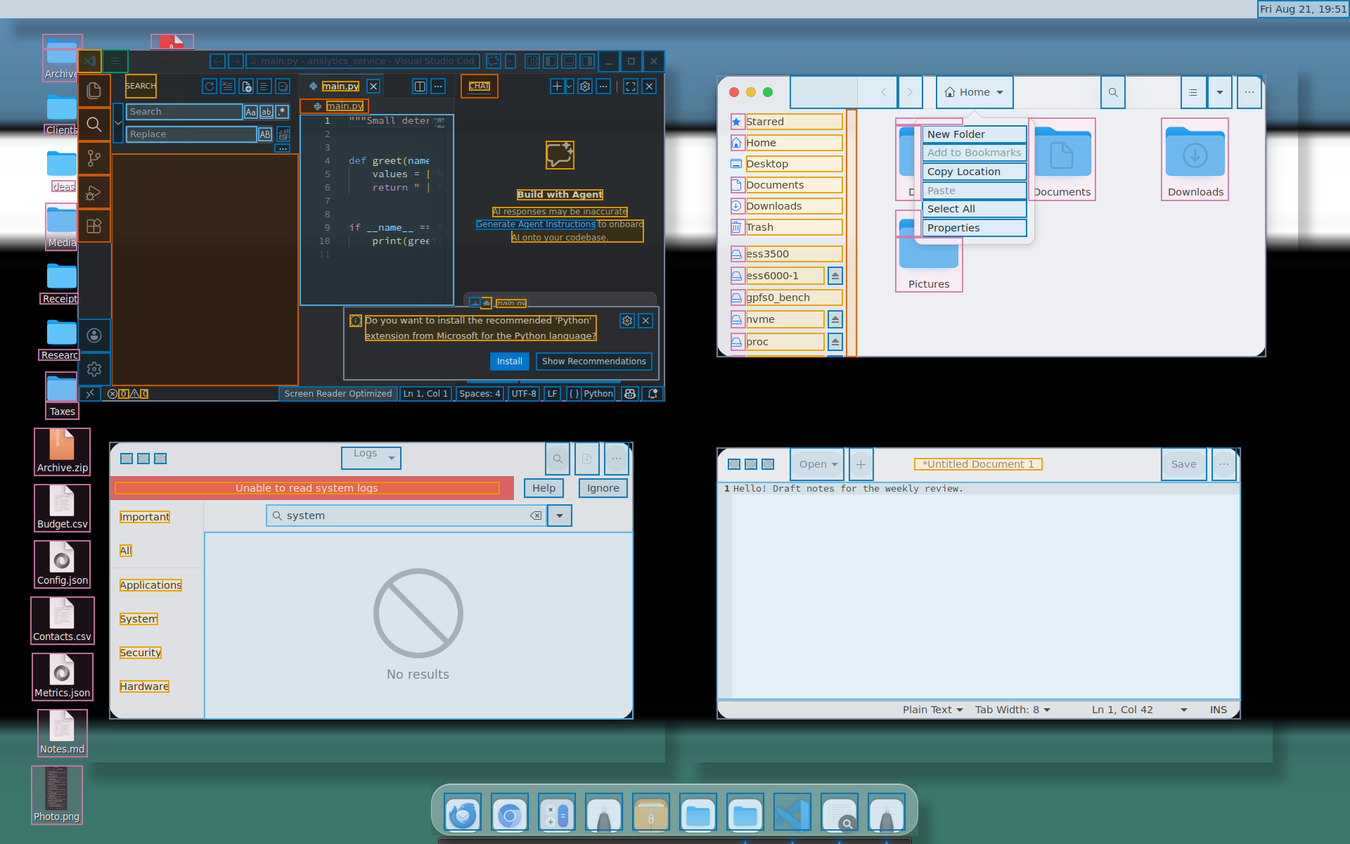} \\
\multicolumn{2}{@{}c@{}}{\scriptsize gnome-logs, gnome-text-editor, nautilus, vscode \textperiodcentered\ 185 annotated elements} \\[0.8mm]
\includegraphics[width=.45\linewidth]{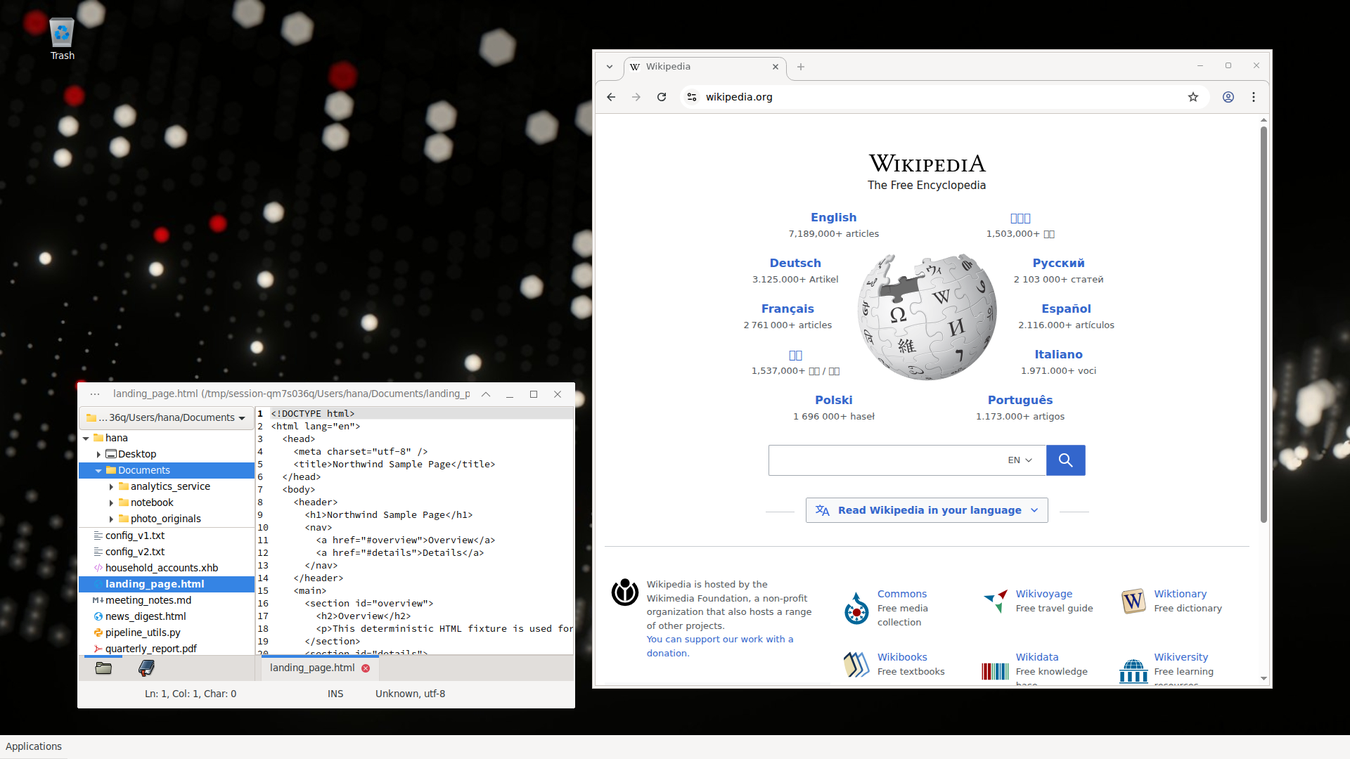} &
\includegraphics[width=.45\linewidth]{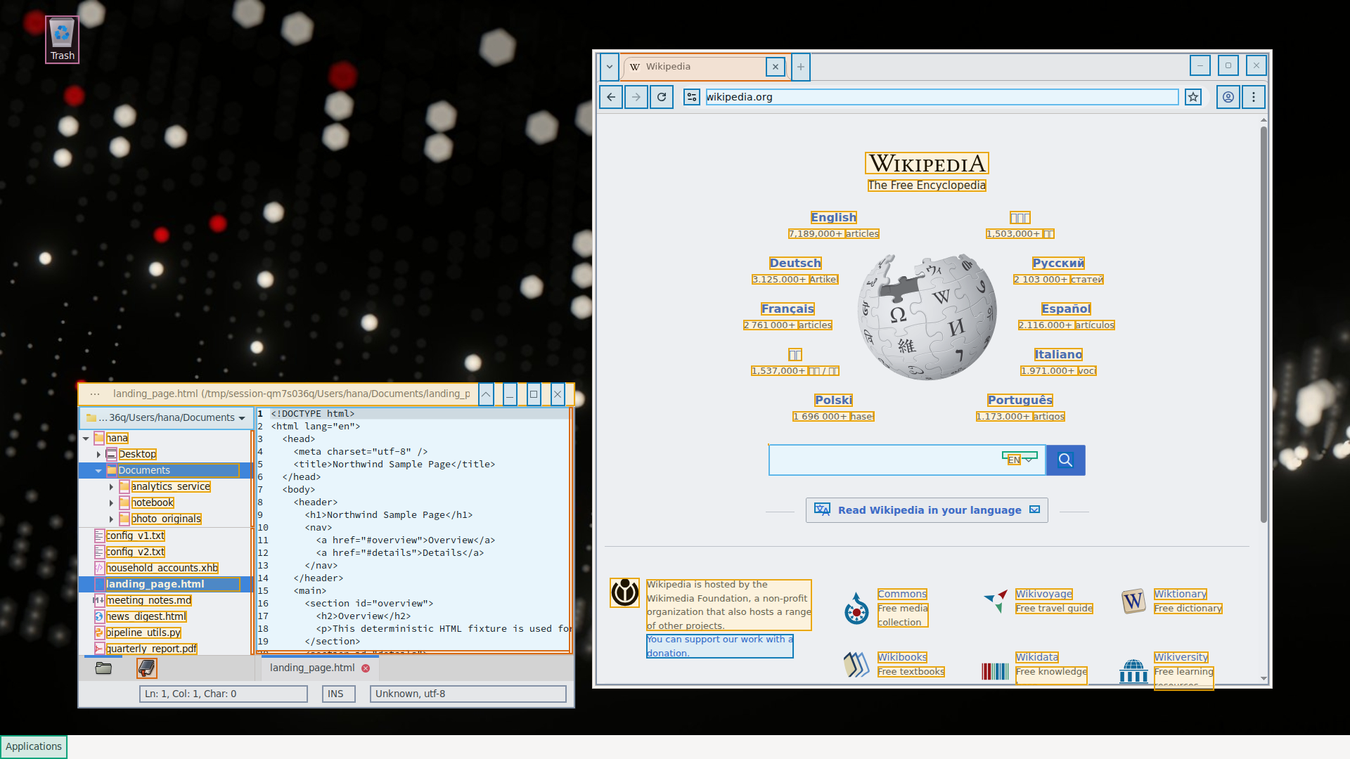} \\
\multicolumn{2}{@{}c@{}}{\scriptsize bluefish, chromium-browser, homebank \textperiodcentered\ 116 annotated elements} \\[0.8mm]
\includegraphics[width=.45\linewidth]{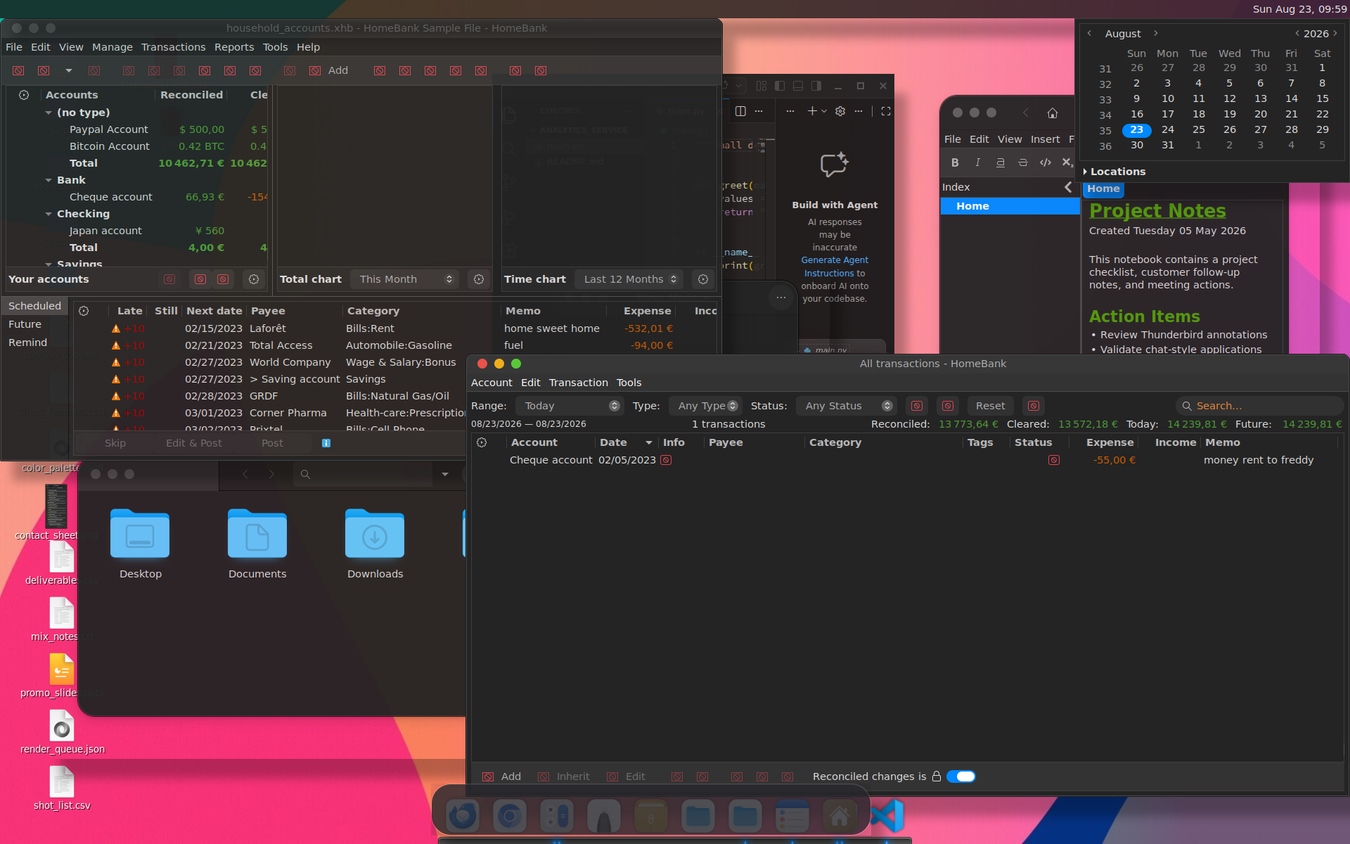} &
\includegraphics[width=.45\linewidth]{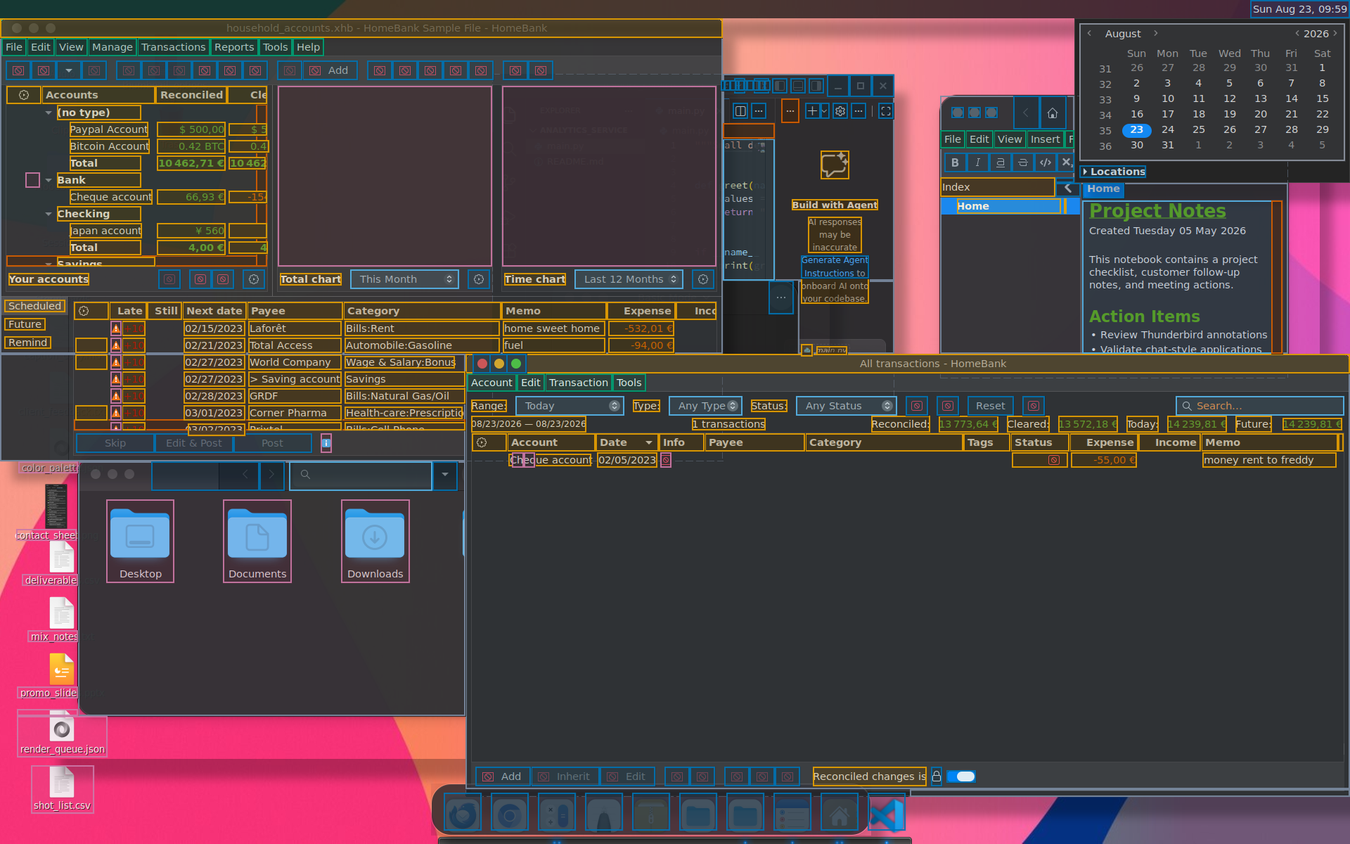} \\
\multicolumn{2}{@{}c@{}}{\scriptsize gnome-calculator, homebank, nautilus, vscode, zim \textperiodcentered\ 271 annotated elements} \\[0.8mm]
\includegraphics[width=.45\linewidth]{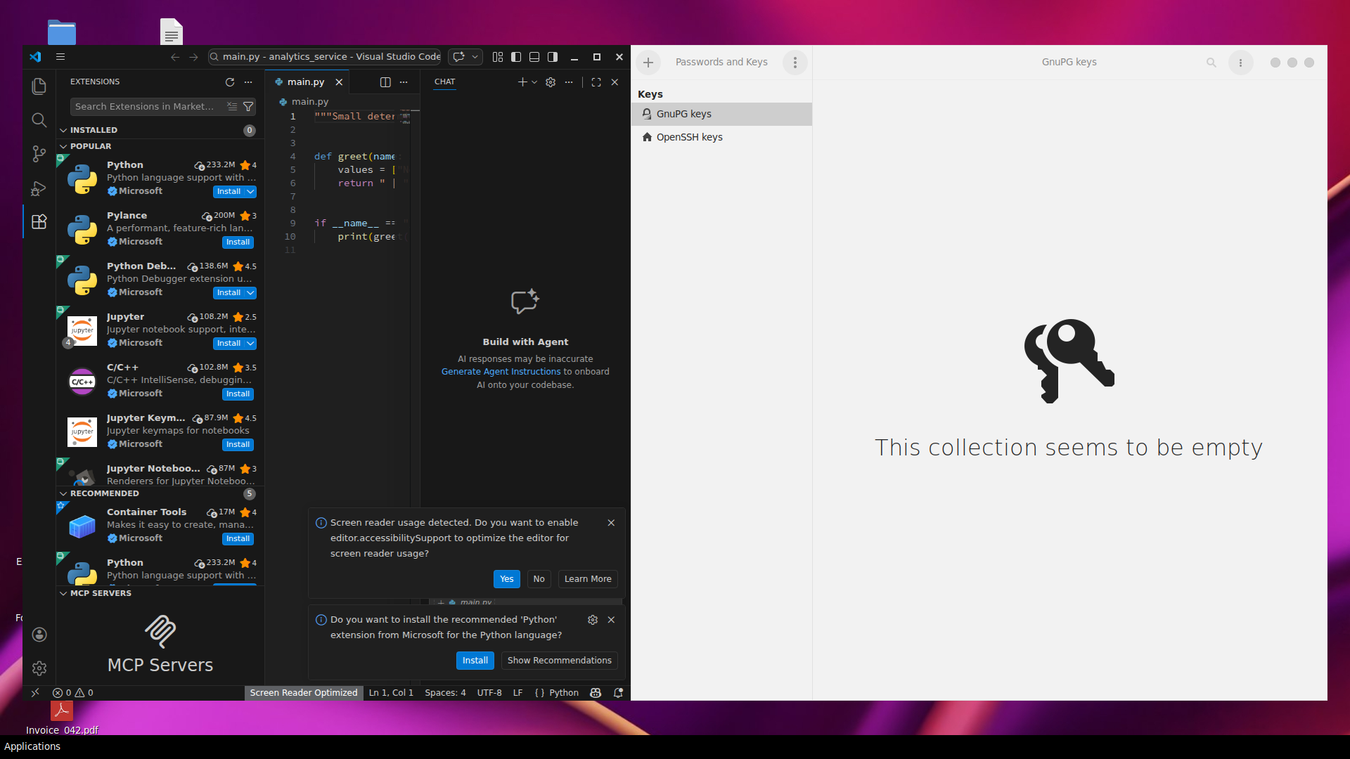} &
\includegraphics[width=.45\linewidth]{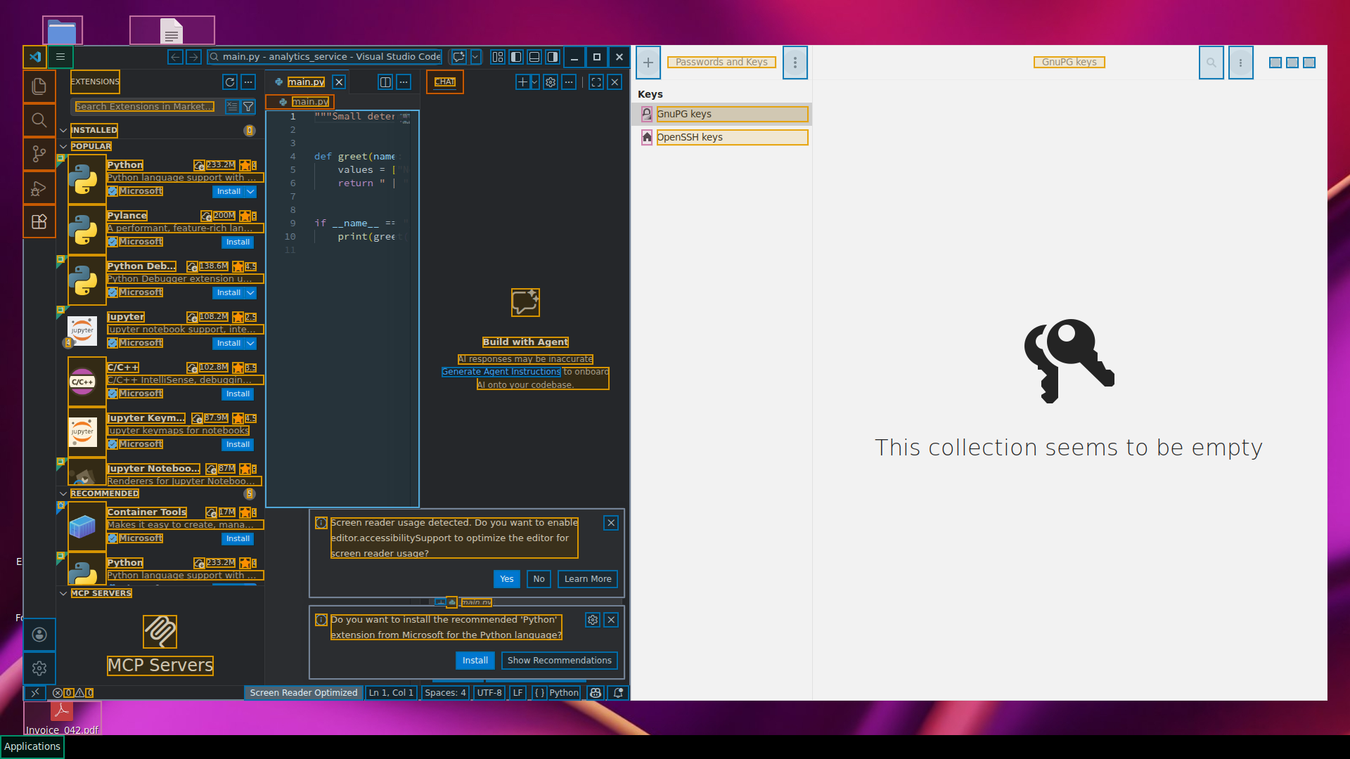} \\
\multicolumn{2}{@{}c@{}}{\scriptsize seahorse, vscode \textperiodcentered\ 201 annotated elements} \\
\end{tabular}
\caption{\textbf{Representative observations and dense annotations.}
Each row pairs a desktop screenshot with its retained element annotations,
colored by element family.}
\label{fig:samples}
\end{figure}

\begin{figure}[t]
\centering
\includegraphics[width=\linewidth]{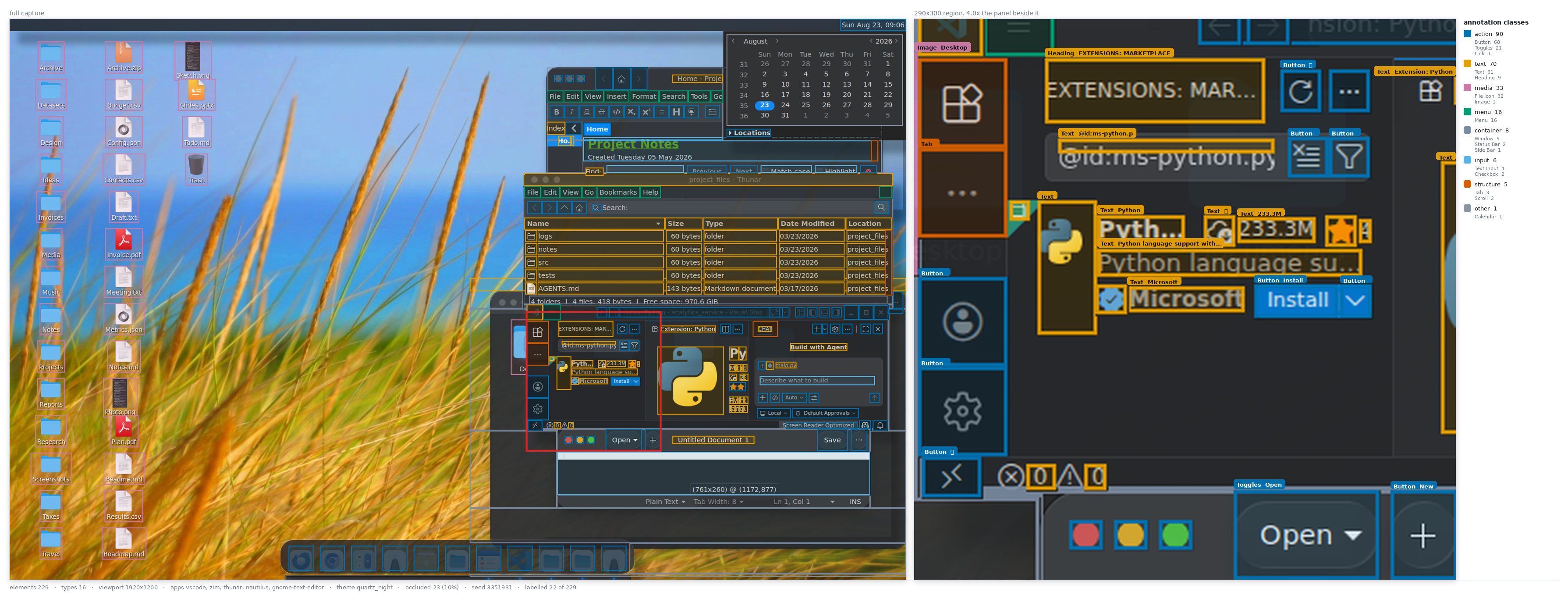}
\caption{\textbf{Detail of a densely annotated region.}
A magnified crop shows retained element geometry and types, with annotations
colored by element family.}
\label{fig:annotationzoom}
\end{figure}

\begin{figure}[t]
\centering
\includegraphics[width=0.68\linewidth]{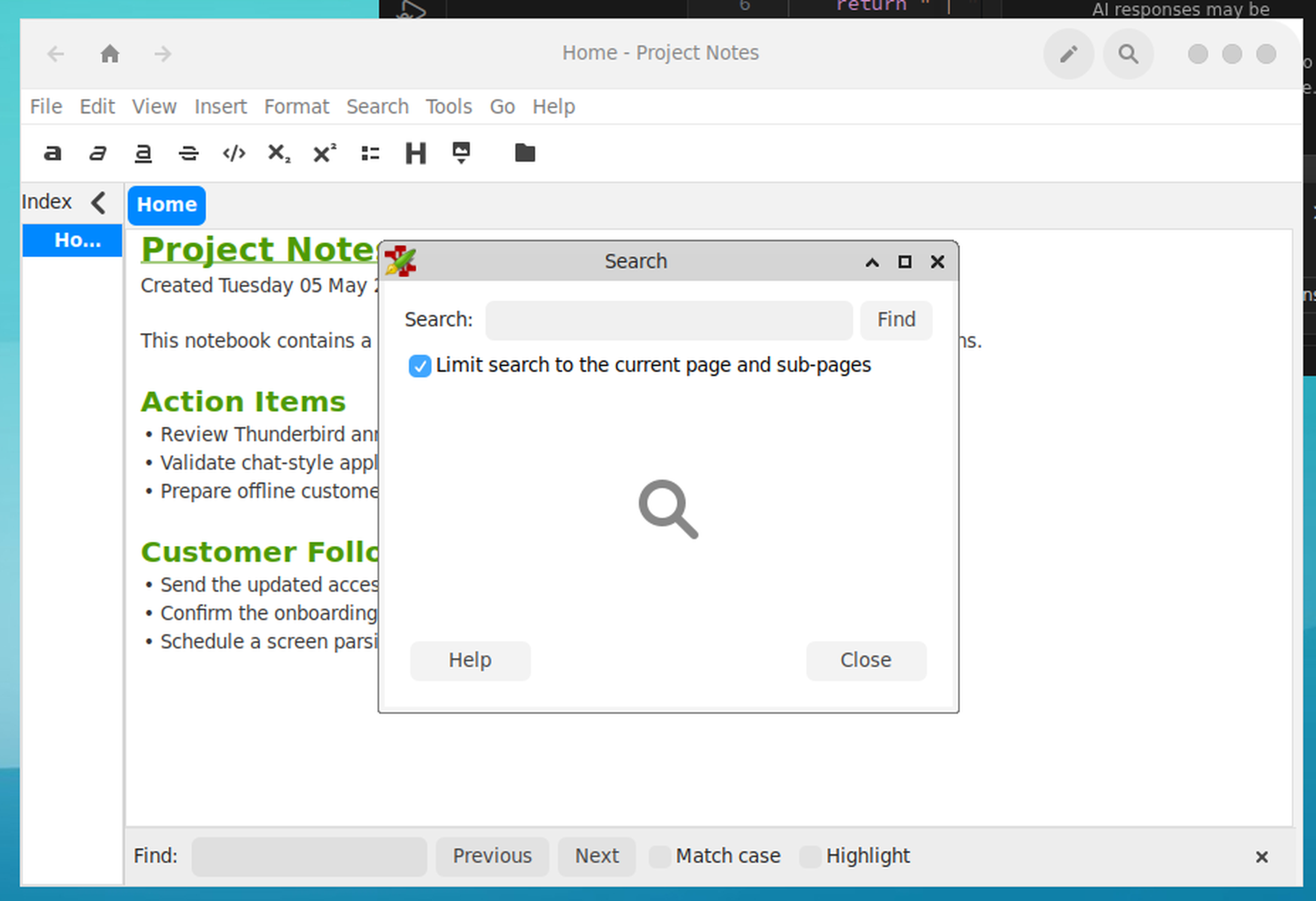}
\par\vspace{2mm}
\begin{minipage}{\linewidth}
\input{figures/screentag_block}
\end{minipage}
\caption{\textbf{An observation and its ScreenTag representation.}
A search dialog partly covers a notebook page. The 26-line serialization excerpt retains
exposed regions as visible fragments and marks occluded portions of text;
coordinates use a 0--500 viewport-normalized grid.}
\label{fig:screentag}
\end{figure}

\begin{figure}[p]
\centering
\includegraphics[width=\linewidth]{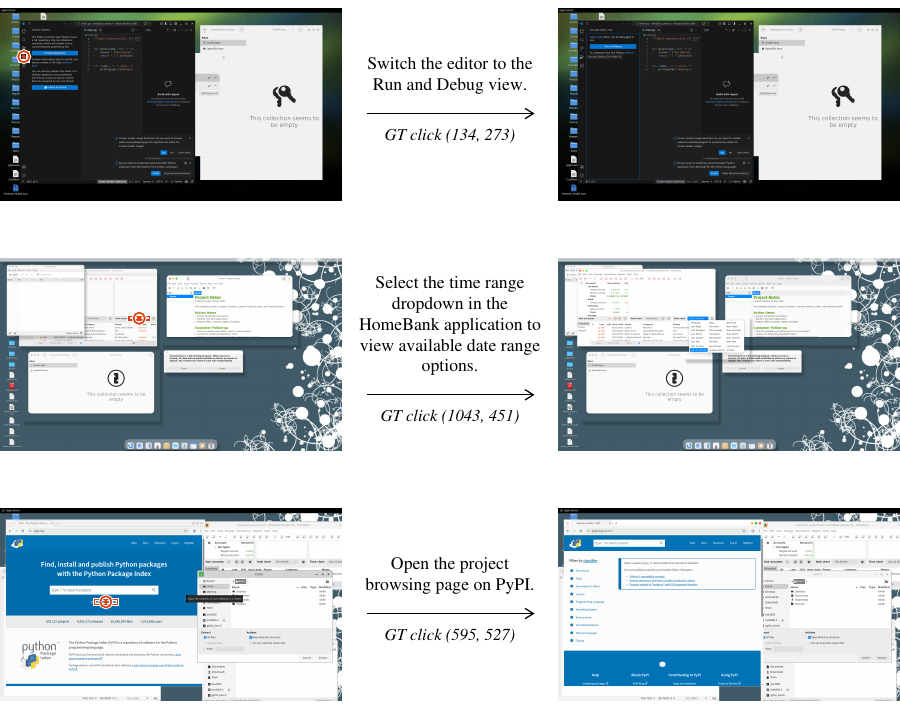}
\caption{\textbf{Recorded interactions and associated instructions.}
Each row shows the initial screen, instruction and recorded click, and
successor screen. Orange circles mark click locations; rectangles show
the recorded target fragment. Coordinates refer to the initial screenshot.}
\label{fig:interactionrecords}
\end{figure}

\end{document}